\documentclass{article} 
\usepackage{iclr2027_conference,times}

\usepackage{amsmath,amsfonts,bm}

\def\eqref#1{equation~\ref{#1}}

\def\1{\bm{1}}

\DeclareMathAlphabet{\mathsfit}{\encodingdefault}{\sfdefault}{m}{sl}
\SetMathAlphabet{\mathsfit}{bold}{\encodingdefault}{\sfdefault}{bx}{n}

\usepackage{hyperref}
\usepackage{url}
\usepackage{graphicx}
\usepackage{booktabs}
\usepackage{multirow}
\usepackage{tcolorbox}
\definecolor{promptbg}{RGB}{245,245,245}
\definecolor{promptframe}{RGB}{60,60,60}
\usepackage{algorithm}
\usepackage{algpseudocode}
\usepackage{amsmath}
\tcbuselibrary{breakable}
\usepackage[accsupp]{axessibility}

\title{LERR-Nav: Learning to Coordinate Exploration, Recovery, and Reminiscing for Object Navigation}

\author{
\textbf{Jingzhi Huang}\textsuperscript{1} \hspace{1.5em}
\textbf{Junkai Huang}\textsuperscript{2} \hspace{1.5em}
\textbf{Yuhang Peng}\textsuperscript{3} \hspace{1.5em}
\textbf{Keyu Mu}\textsuperscript{4} \hspace{1.5em}
\textbf{Hailong Huang}\textsuperscript{1} \and
\textbf{Haoyang Yang}\textsuperscript{1} \hspace{1.5em}
\textbf{Renbo Luo}\textsuperscript{5} \hspace{1.5em}
\textbf{Lujia Wang}\textsuperscript{4} \hspace{1.5em}
\textbf{Yi Wang}\textsuperscript{1}\textsuperscript{\dag} \\[1.5ex]
\textsuperscript{1}Hong Kong Polytechnic University \quad
\textsuperscript{2}Institute of Automation, Chinese Academy of Sciences \\
\textsuperscript{3}DeepRobotics \quad
\textsuperscript{4}Hong Kong University of Science and Technology (Guangzhou) \\
\textsuperscript{5}Guangzhou University \\[1ex]
\texttt{jingzhi.huang@connect.polyu.hk}, \texttt{yi-eie.wang@polyu.edu.hk} \\
\textsuperscript{\dag}Corresponding author \\
\url{https://github.com/kim-os/LERR-Nav}
}

\iclrfinalcopy 
\begin{document}

\maketitle
\fancyhead{} 
\renewcommand{\headrulewidth}{0pt} 
\fancyfoot[C]{\thepage} 

\begin{abstract}
Object Navigation in unseen environments remains a fundamental challenge in embodied AI, as it requires agents to generalize visual and spatial reasoning to novel layouts. While recent advances in large language model (LLM) and vision-language model (VLM) have enhanced semantic mapping, spatial memory, and waypoint selecting for  navigation, successful navigation requires more than selecting the next promising waypoint. A robot must also recognize stalled progress, escape unproductive behaviors, and revisit overlooked target cues. Addressing these challenges requires both specialized navigation strategies and a high-level policy that coordinates their execution. Motivated by the analysis of failed navigation episodes, we propose LERR-Nav, an object navigation framework that coordinates exploration, recovery, and reminiscing through a learned high-level policy. The framework comprises four complementary navigation strategies: value-function-guided exploration, LLM-guided exploration, recovery from stalled or repetitive behavior, and memory-guided reminiscing. To support history-aware coordination, we maintain the execution memory that preserves textual summaries of all strategy executions alongside a sliding window of map sequences. To learn the coordination policy, we initialize a high-level vision-language agent from Qwen2.5-VL-7B-Instruct and train it through a two-stage curriculum. The first stage uses map-centric supervised fine-tuning (SFT) to establish spatial grounding. The second stage performs SFT on variable-length navigation segments to learn which navigation strategy to invoke next based on current maps and the hybrid execution memory. Experiments on HM3D and MP3D demonstrate state-of-the-art navigation performance, while ablation studies validate the contributions of individual components. We further deploy our LERR-Nav on a Unitree Go2 quadruped robot for real-world object navigation.
\end{abstract}

\section{Introduction}
\label{sec:intro}
Object Navigation is a fundamental and challenging task in Embodied AI, requiring agents to locate specific objects in an unknown, complex environment based on a natural language description of the target category and visual information \cite{Obj-task}. Compared to conventional data-driven learning methods\cite{SemExp,train1,RIM,XGX,OVG-Nav,SEEK,RL-OGN}, the zero-shot paradigm\cite{Openfmnav,Cl-cotnav,SemNav,Vln-game,Loc-zson,L-ZSON,Esc} can adapt to unseen environments without the need for additional training data. In practical scenarios, for example the open-world environments, this capability allows robots to navigate a completely new environment to retrieve specific items (e.g., a book) for patients. Furthermore, from the perspective of real-world deployment.

Early  methods are primarily designed for single-floor environments. Initially, \cite{Cows,embeddings} leveraged Vision-Language Model (VLM) to extract semantic features from RGB observations, which were then integrated with exploration strategies to accomplish object-goal navigation tasks. However, these  approaches proved highly inefficient. With the rapid advancement of Multimodal Large Language Model (LLM), recent works seek to leverage the powerful reasoning capabilities of LLM to improve navigation efficiency. Consequently, LLM are frequently employed to select waypoints based on current RGB image\cite{Curiosity,Instructnav,Imaginenav}, value map\cite{HuLE-Nav,Openfmnav}. Nevertheless, these methods struggle to generalize to multi-floor navigation tasks. As the scale of the environment expands, the demands on the agent for complex geometric perception and path planning increase significantly.

Recent object navigation studies recognize this limitation and propose solutions for navigating across multiple floors. Spatially, methods such as \cite{SpatialNav,3DGSNav,3D-mem} utilize 3D Simultaneous Localization and Mapping (SLAM) to build maps and \cite{MFNP,Sg-nav,Open_scene} construct hierarchical topology maps from floors to rooms. To improve efficiency, approaches like \cite{Stairway} prioritize searching nearest frontiers and rely on LLM for distant planning. Additionally, methods like \cite{exp_pro1,Beliefmapnav} focus on finding the shortest exploration paths, method \cite{Cognav} further decomposes the exploration strategy to emulate human cognition, it does so at the cost of making each decision step heavily dependent on an LLM, leading to substantial token consumption.
Despite improving object-finding success rates by enhancing spatial representations and boosting exploration efficiency, these works overlook two critical issues.
First, their frameworks lack a well-defined strategy representation and strategy transitions; they often overemphasize exploration, leaving robots less robust during navigation and unable to cope effectively with abnormal situations. For example, the robot may fail to locate the entrance to a staircase or get stuck while attempting to reach a frontier point.
Second, their exploration strategies fail to adequately adapt to environmental variations in exploration coverage, semantic cues, and structural information and so on, making it difficult to balance exploration, exploitation, and high-level planning.

In summary, the contributions of our method are as follows: (1) Motivated by an analysis of failed navigation episodes,
we formulate robust object navigation beyond waypoint selection
as the coordination of exploration, recovery, and reminiscing.
We instantiate this formulation with four complementary
strategies: value-function-guided exploration, LLM-guided
exploration, recovery from stalled or repetitive behavior,
and memory-guided reminiscing.
(2) We introduce a learned high-level vision-language policy
to coordinate these strategies at their execution boundaries.
The policy is supported by an execution memory that combines
persistent textual summaries with time-sparsified visual
history, and is trained through a two-stage SFT curriculum
that progresses from map-centric spatial grounding to
history-aware strategy coordination.
(3) LERR-Nav achieves state-of-the-art performance on HM3D
and MP3D, improving SR over the strongest baselines by
6.9 and 3.4 percentage points, respectively.
Extensive ablations validate the contributions of the
navigation strategies, execution memory, and coordination
policy, while deployment on a Unitree Go2 quadruped
demonstrates the applicability of LERR-Nav to real-world
object navigation.

\section{Related Works}
\textbf{Object Navigation.} Object navigation refers to the task of guiding a robot to search for a given target object within an unknown environment. The dominant strategies can generally be divided into two primary categories: the first group includes training-dependent methods, such as reinforcement learning and imitation learning, which necessitate extensive training using task-specific datasets\cite{train1, train2, train3, train4, train5}. The second category includes zero-shot methods\cite{Openfmnav,Cl-cotnav,SemNav,Vln-game,Loc-zson,L-ZSON,Esc}, which leverage pre-trained models like VLM\cite{Blip-2,Clip} or LLM to enable the robot to exhibit strong adaptability in unseen environments. These methods commonly utilize LLM as decision-maker. However, they often neglect to consider the robot's strategy. Therefore, our zero-shot approach incorporates an LLM role-playing strategy, allowing the robot to take different actions depending on its current strategy, thus improving the decision-making process in dynamic and varied contexts. \\

\noindent \textbf{Exploration Strategy for  Object Navigation.} Under a strict step budget, balancing exploration and exploitation in complex multi-floor environments is crucial for robots. Most early methods rely heavily on greedy strategies. SSR-ZS\cite{SSR-ZS} constructs semantic maps and utilizes LLM to identify and select target-relevant frontiers; VLFM\cite{Vlfm} employs VLM to build value maps linking frontiers to the target; and SGNav\cite{Sg-nav} leverages 3D scene graphs to prompt LLM for spatial reasoning. However, strategies that depend so heavily on semantic or LLM-derived scoring often suffer from inefficient exploration.
Subsequent approaches have sought to address this limitation. ApexNav\cite{Apexnav} introduces an adaptive exploration method that dynamically switches operational modes based on the strength of semantic cues. However, the lack of continuous LLM reasoning during the exploration process often leads the robot to overlook areas where the target is hidden, particularly when the semantic cues are weak. Similarly, while ASC\cite{Stairway} proposes a coarse-to-fine exploration strategy that combines nearest-frontier exploration with LLM guidance, it fails to adapt dynamically to environmental variations, causing the robot to simply navigate to the nearest frontier most of the time. In short, we should use the LLM in moderation and integrate it into the exploration strategy. So, we propose an environment-adaptive exploration strategy integrated with a fast-and-slow thinking mechanism.
\section{Method}
\begin{figure}[t]
    \centering
    \includegraphics[width=0.85\textwidth]{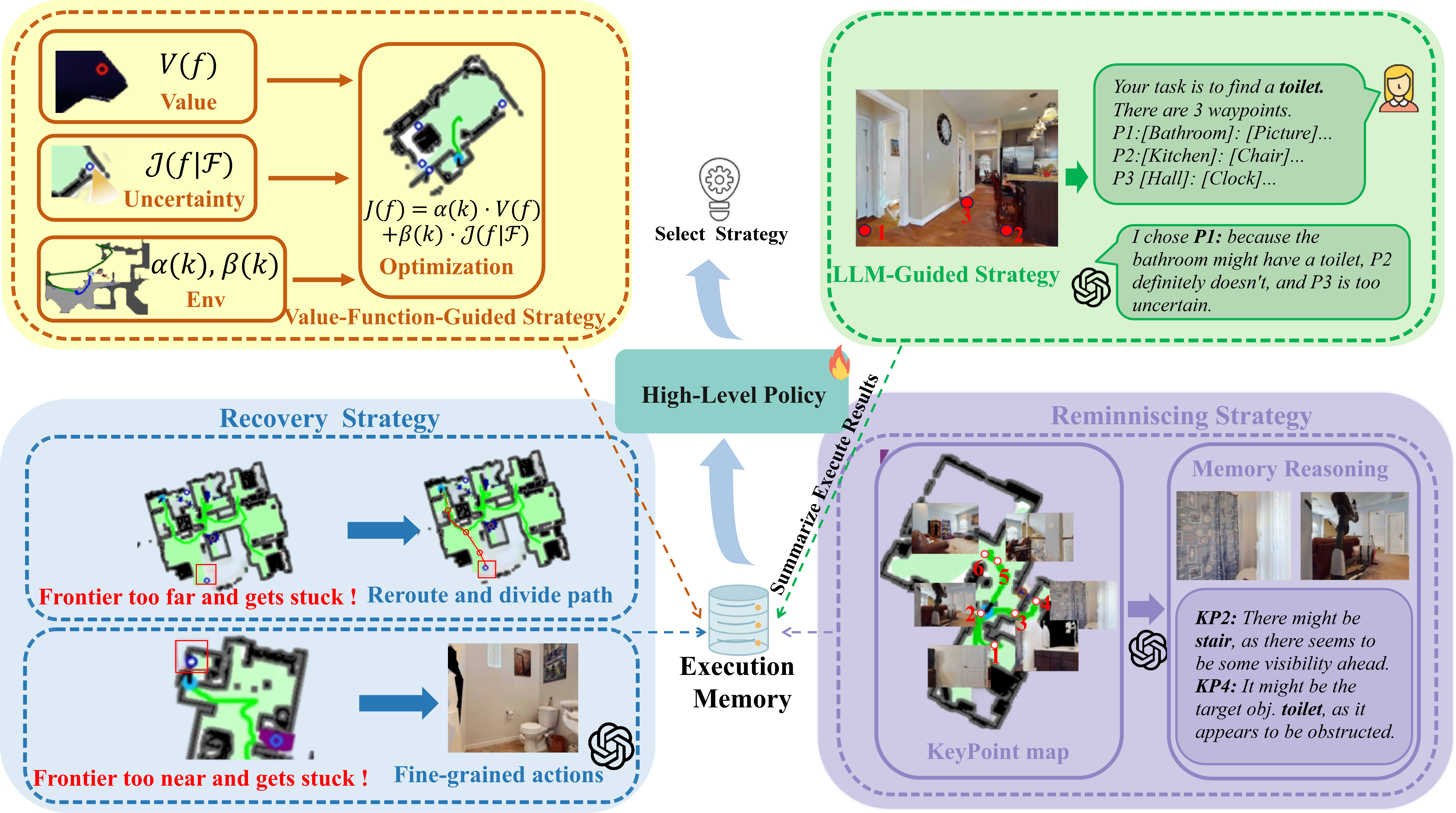}
    \caption{\textbf{LERR-Nav pipeline:}
Conditioned on the target instruction and execution memory,
a learned high-level policy invokes one of four complementary
strategies: value-function-guided exploration, LLM-guided
exploration, recovery, or reminiscing.
The selected strategy interacts with the environment until
termination, after which its textual execution summary and
RGB--map records are used to update the execution memory.
The updated memory then informs the next high-level
strategy invocation, forming a closed-loop navigation process.
    }
    \label{fig:1}
\end{figure}

\subsection{Task Definition}
We formulate Object Navigation task, the robot needs to locate target objects in unfamiliar environments it has never encountered before. The target object $c$ is provided as a free-form linguistic description. At each discrete time step $t$, the agent perceives the environment through an RGB-D sensor. Concurrently, the agent tracks its relative strategy $s_t = (x_t, r_t)$ via on-board odometry, representing its position $x_t \in \mathbb{R}^3$ and orientation $r_t \in SO(3)$. The agent's decision-making framework must select a discrete action $a_t$ from the action space $\mathcal{A}$: = \{\text{MOVE\_FORWARD}, \text{TURN\_LEFT}, \text{TURN\_RIGHT}, \text{LOOK\_UP}, \text{LOOK\_DOWN}, \text{STOP}\}. The locomotion is executed in fixed increments (e.g., $0.25$m for translation and $30^\circ$ for rotation). An episode is deemed a success if the agent issues a $\text{STOP}$ command within a proximity threshold of $0.1$m from the target object, without exceeding a maximum budget of $500$ steps.
\subsection{Framework Overview}

As illustrated in Fig.~\ref{fig:1}, LERR-Nav is an object navigation
framework that coordinates exploration, recovery, and
reminiscing through a learned high-level policy.
During navigation, the robot maintains different maps to support
 reasoning and preserve informative observations
(Sec.~\ref{sec:Map Representation}). Building on these representations, the framework provides
four complementary navigation strategies, as described in Sec.~\ref{sec:Complementary Navigation Strategies}.  The execution memory (Sec.~~\ref{sec:Execution Memory with Time Sparsification}) offer agent Long-Term Memory. A learned high-level policy coordinates strategy execution
and is trained through two-stage SFT (Sec.~\ref{sec:High-Level Policy}).

\subsection{Map Representation}
\label{sec:Map Representation}
\textbf{Visibility Map} $\mathcal{M_\text{vis}}$. We maintain a 2D occupancy grid map  by projecting depth observations into 3D point cloud and subsequently performing Bird's Eye View projection. Based on the boundaries between explored free space and unexplored regions, we extract a set of candidate frontiers $\mathcal{F} = \{f_1, f_2, \dots, f_n\}$. These frontiers are categorized into two subsets: intra-floor frontiers $\mathcal{F}_\text{floor}$ and staircase frontiers $\mathcal{F}_\text{stair}$, where staircases are treated as traversable transitions. The specific geometry-based staircase detection and frontier generation logic follow the \cite{Stairway}.\\

\noindent \textbf{Exploration Value Map} $\mathcal{M_\text{val}}$.  We define a value  function $V(f)$ for each frontier $f \in \mathcal{F_\text{floor}}$ as follows:
$
\label{eq1}
V(f) = \alpha \cdot S_\text{sem}(f) + \beta \cdot S_\text{dist}(f),
$
where the $S_{\text{sem}}(f)$ evaluates the semantic relevance of a frontier. Using the BLIP-2\cite{Blip-2}, we can calculate the score by measuring the cosine similarity between an RGB observation $I_f$ at the frontier and a set of context-aware text prompts $T$ (e.g., bathtub or sink when the target is toilet) generated by the LLM's common-sense reasoning. $S_{\text{dist}}(f)$ favors proximal exploration by employing a linear decay function based on the geodesic distance $d(r, f)$ between the robot's current position $r$ and the frontier $f$:
$
S_{\text{dist}}(f) = \max\left(0, 1 - \frac{d(r, f)}{d_{\text{max}}}\right),
$ where $d_{\text{max}}$ represents  a predefined normalization constant.\\

\noindent \textbf{KeyPoint Map} $\mathcal{M_\text{key}}$. Inspired by the keyframes of Orb-SLAM2\cite{Orb-slam2}, the difference is that these keyframes are fixed onto a 2D map as keypoints. The criteria for selecting keypoints are as follows: all room entrances and frontiers with open visibility. Frontiers with open visibility can be quantified by the area of the robot's field of view (FOV) at that point, which is not obstructed by obstacles.

We independently maintain these three distinct map layers for each floor, resulting in a global environmental representation structured as a set of floor-specific tuples: $\mathcal{M} = \{ \langle F_i, \mathcal{M}_{\text{vis}(i)}, \mathcal{M}_{\text{val}(i)}, \mathcal{M}_{\text{key}(i)} \rangle \}_{i=1}^{n}$, where $F_i$ denotes the $i$-th floor index.

\subsection{Complementary Navigation Strategies}
\label{sec:Complementary Navigation Strategies}
\subsubsection{Value-Function-Guided Exploration Strategy}
\label{sec:3.5}
We formulate the frontier selection as an optimization problem that balances current values priors with the expected reduction in environmental uncertainty. Let $\mathcal{Z}_k$ be the observations up to step $k$, and $\rho_{k|k} = p(\Theta | \mathcal{Z}_k)$ denote the current belief of the target $\Theta$. When the robot evaluates a candidate frontier $f \in \mathcal{F}_{\text{floor}}$, it anticipates a future belief $\hat{\rho}_{k+1|k} = p(\Theta | \mathcal{Z}_k, \hat{z}_{k+1}) = {\mathcal{I}(f | \mathcal{F})}$, where $\hat{z}_{k+1}$ is a virtual estimate within the FOV at $f$:
$
\label{eq:objective}J(f) = \alpha(k) \cdot V(f) + \beta(k) \cdot {\mathcal{I}(f | \mathcal{F})},
$
where $\alpha(k)$ and $\beta(k)$ are factors that change with the environment, and will be explained later, $V(f)$ is the prior value, and $\mathcal{I}(f | \mathcal{F})$ quantifies the expected uncertainty reduction. 
To compute this, we first project the FOV at $f$ to obtain the coverage area $S_1$. For the set of semantic scores $\mathcal{C}$ at the boundaries of $S_1$, we apply a Gaussian kernel $G(\cdot)$ to estimate the spatial information density $d(\mathbf{p}) = G(\mathcal{C})$. \\ 
To account for the redundancy and overlap between multiple frontiers, the total estimated uncertainty $f_{1\_u}$ for a frontier is adjusted by considering the mutual information between overlapping coverage areas. Specifically, the integrated uncertainty at $f$ is formulated as: $\mathcal{I}(f | \mathcal{F}) = \int_{S_1(f)} d(\mathbf{p}) d\mathbf{p} + \sum_{f_j \in \mathcal{F}, j \neq f} \text{Area}\left( S_1(f) \cap S_1(f_j) \right),
$ where $\text{Area}(S_1(f) \cap S_1(f_j))$ represents the spatial overlap between the estimated coverage of frontier $f$ and its neighbors $f_j$, more details can be found in the Appendix~\ref{sec:Value-Function-Guided Exploration Strategy  Details}. 

\(\alpha(k)\) and \(\beta(k)\) are adapted based on the Exploration Reward (ER), which characterizes the robot's exploration progress and the remaining environmental potential at time $k$. The exploration reward $\text{ER}(k)$ is defined as a multi-factor metric:
\begin{equation}\text{ER}(k) = \sigma_1 \cdot \frac{|U(k)|}{|E|} + \sigma_2 \cdot \frac{N_{\text{frontier}}(k)}{N_{\text{total}}} + \sigma_3 \cdot \left(1 - \frac{k}{K_{\text{max}}}\right),\end{equation}
where $|U(k)|/|E|$ represents the ratio of the unexplored area to the total environment size at step $k$, $N_{\text{frontier}}(k)/N_{\text{total}}$ denotes the current density of discovered frontiers, and the final term introduces a temporal decay to prioritize rapid exploration in the early stages of the mission. The strategy-dependent coefficients $\alpha(k)$ and $\beta(k)$ are updated as follows: $\beta(k) = \beta_{\text{max}} \cdot \text{ER}(k),  \alpha(k) = \alpha_{\text{min}} \cdot (1 - \text{ER}(k)).$ 

\subsubsection{LLM-Guided Exploration Strategy}
A local room and object topology map is first constructed to enhance the accuracy of LLM predictions. First, the current RGB image is divided based on the regions of doorframes. Then, each RGB image is processed using RAM\cite{ram} to classify the object categories, and Place365\cite{Places} is used to identify the room types, ultimately generating descriptions for different rooms. Following this, prompts are designed to allow LLM to evaluate, in conjunction with the visibility map and the current RGB image, whether a frontier point is suitable for exploration.

\subsubsection{Recovery Strategy}
\label{sec:3.6}
As shown in Fig. \ref{fig:4}(b), when dealing with distant frontier point, the A* algorithm is first applied on the 2D visibility map to compute a collision-free path from the current position to the target. By employing an interval sampling method, several intermediate points are generated along the path. The target, therefore, becomes to traverse all the intermediate points, and after testing, the robot successfully resumes normal movement.

As shown in Fig. \ref{fig:4}(e), when handling nearby frontier points, LLM is used to generate fine-grained action commands by integrating the RGB image and the local map.

\subsubsection{Reminiscing Strategy}
\label{sec:3.7}
Following the completion of single-floor exploration, a two-stage reminiscing process is initiated to ensure thoroughness and facilitate multi-floor transitions. This process relies on $\mathcal{M_\text{key}}$, where each point preserves a corresponding RGB observation  on the global map. This integration allows the LLM to perform joint spatial-semantic reasoning by synthesizing the global geometric structure with localized visual data. As shown in Fig. \ref{fig:4}, the reminiscing logic is executed in two hierarchical stages:

\textbf{Target Verification}. The LLM first reviews the aggregate keypoint collection to identify potential nodes where the target object may have been overlooked. Upon selecting candidate keypoints, the system extracts the RGB images for a secondary, fine-grained confirmation. If verified, the robot navigates to the specific coordinates and executes a terminal approach to the target.

\textbf{Look for Staircase}. If no potential target locations are identified, the LLM shifts focus to detecting potential staircase entrances to enable vertical mobility. The robot employs the PointNav~\cite{wijmans2022ver} policy to navigate toward the keypoint with the highest probability of containing a staircase. Final alignment with the staircase entrance is achieved through LLM-guided fine-grained action.

\subsection{Execution Memory with Time Sparsification.}
\label{sec:Execution Memory with Time Sparsification}
We maintain an
execution memory that combines textual summaries of all
completed strategy executions with a sliding window of
RGB observations and map snapshots. 
Let $t_k$ denote the environment timestep of the $k$-th
high-level decision.
Execution segment $j$ begins when a strategy is invoked
at $t_j$ and ends when it terminates at $t_{j+1}$.
For each completed segment, we record a textual summary
$e_j$ and a temporally ordered sequence of RGB--map pairs:
$
\mathbf{V}_j
=
\bigl(
(t,I_t,M_t)
\bigr)_{t\in\mathcal{T}_j},
$ where $\mathcal{T}_j$ contains the recorded timesteps
within segment $j$, $I_t$ is the RGB observation, and
$M_t$ is the corresponding rendered navigation map.
The summary $e_j$ describes the executed strategy, its
observed outcome, the start and end locations, and the
rooms or landmarks encountered during execution.

At decision $k$, the visual window contains at most
the $W$ most recent completed execution segments:
$
\mathcal{W}_k
=
\left\{
j \mid \max(1,k-W)\leq j\leq k-1
\right\}.
$
We apply time sparsification  to the visual records
within this window.
Specifically, we measure the age of segment $j$ by its
distance from the current decision, $a=k-j$, and assign
it a non-increasing visual retention budget $b(a)$:
$
b(a+1)\leq b(a),
\left|
\widetilde{\mathbf{V}}_j^{(k)}
\right|
\leq
\min\left\{
|\mathbf{V}_j|,\,b(k-j)
\right\},
$ where $\widetilde{\mathbf{V}}_j^{(k)}$ is the retained
visual subsequence of segment $j$ at decision $k$.
The retained RGB observations and map snapshots remain
paired, with their original timestamps and chronological
order preserved.
Time sparsification therefore reduces the number of
historical snapshots rather than their spatial resolution.

The resulting execution memory is
$
\mathcal{E}_k
=
\left(\mathcal{H}_k,\mathcal{V}_k\right),
\mathcal{H}_k
=
(e_1,\ldots,e_{k-1}),
\mathcal{V}_k
=
\left(
\left(j,\widetilde{\mathbf{V}}_j^{(k)}\right)
\right)_{j\in\mathcal{W}_k}.
$ As the window advances, visual records outside
$\mathcal{W}_k$ are removed from the coordinator's visual
memory, while their textual summaries remain in
$\mathcal{H}_k$.

\subsection{High-Level Policy}
\label{sec:High-Level Policy}
\begin{figure}[t]
    \centering
    \includegraphics[width=\textwidth]{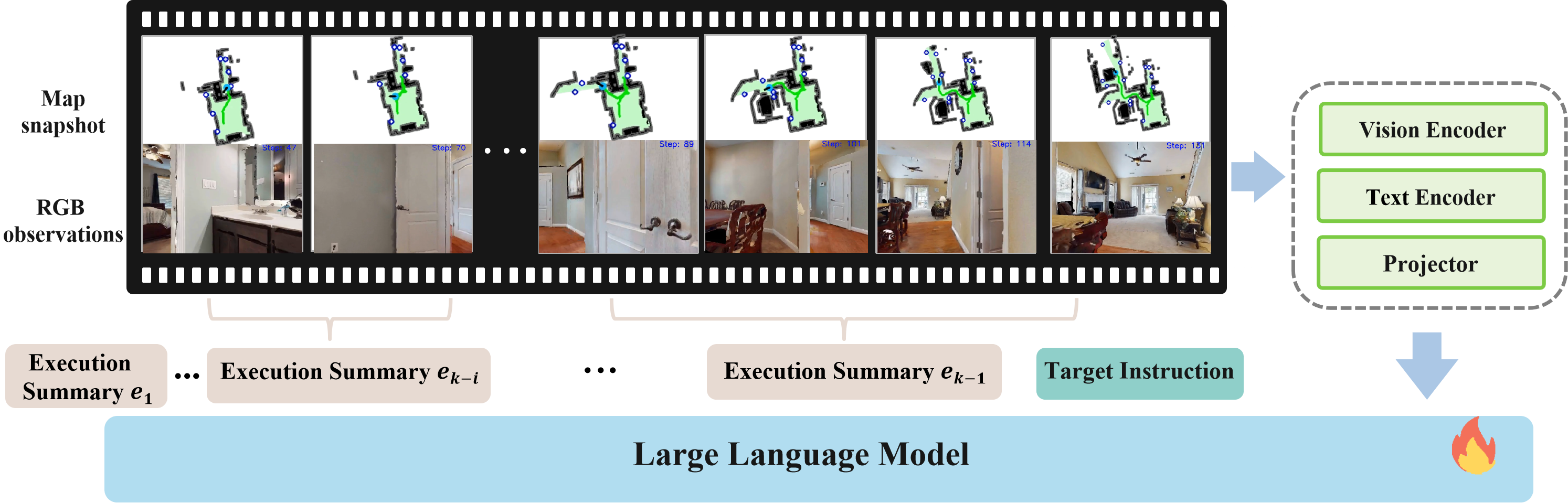}
    \caption{\textbf{Pipeline of the high-level coordination policy.}
At the $k$-th decision, execution memory offers
textual summaries of all completed strategy executions
($e_1,\ldots,e_{k-1}$), and historical RGB
observations and map snapshots to the agent. }
    \label{fig:policy_pipeline}
\end{figure}

We instantiate the high-level coordination policy with
Qwen2.5-VL-7B-Instruct~\cite{bai2025qwen25vl}.
Rather than predicting low-level action, the agent
determines which navigation strategy to invoke after the
active strategy terminates, as shown in Fig.~\ref{fig:policy_pipeline}. 

\noindent\textbf{Stage I: Map-Centric Spatial and Trajectory
Understanding.}
The first stage adapts the model to the map representation
used by the navigation system through map-centric visual
question answering (VQA).
We construct a task-specific VQA dataset from rendered
navigation maps and their associated execution records:
$
\mathcal{D}_{\mathrm{map}}
=
\left\{
(\mathbf{M},q,y)
\right\},
$ where
$\mathbf{M}=(M_{\tau_1},\ldots,M_{\tau_L})$
is a temporally ordered sequence of map snapshots,
$q$ is a map-related question, and $y$ is the reference
answer.

The questions cover three complementary aspects:
\textit{spatial configuration}, including the number
and relative distribution of marked frontier candidates;
\textit{map evolution}, including changes in observed
regions across snapshots; and
\textit{trajectory interpretation}, including movement
direction, displacement, and repeated traversal of
previously visited areas.

We optimize a unified answer-generation objective:
$
\mathcal{L}_{\mathrm{map}}(\theta)
=
-\mathbb{E}_{\mathcal{D}_{\mathrm{map}}}
\Bigl[
|y|^{-1}
\sum_{j=1}^{|y|}
\log P_{\theta}(y_j \mid \mathbf{M},q,y_{<j})
\Bigr].
$
where the expectation is taken over
$(\mathbf{M},q,y)\sim\mathcal{D}_{\mathrm{map}}$, $\theta$ denotes the trainable model parameters
and $y_{<j}$ denotes the preceding answer tokens.
Details of dataset construction, question templates,
answer annotation rules, and representative VQA examples
are provided in Appendix~\ref{app:map_vqa}.

\noindent\textbf{Stage II: Supervised Strategy Coordination.}
Starting from the Stage I checkpoint, we further adapt the
model to coordinate navigation strategies through supervised
demonstrations. The training inputs follow the same organization as policy inference.

We collect navigation rollouts under the high-level guidance
of GPT-6 Astra~\cite{openai2026gpt6astra}, which determines which navigation strategy
to invoke at each decision boundary.
Rollouts that achieve task success are retained and subjected
to manual quality screening.
An LLM-based annotation tool then organizes the retained
demonstrations into strategy-execution segments, with segment
boundaries aligned with the recorded strategy invocations
and terminations.

For each recorded decision $k$, we reconstruct the execution
memory using only the preceding completed segments.
The input is
$\mathbf{x}_k =
\operatorname{Pack}(g,\mathcal{H}_k,\mathcal{V}_k)$,
where $\mathcal{H}_k$ contains the preceding textual execution
summaries and $\mathcal{V}_k$ is constructed using the same
sliding-window and time-sparsification mechanism as inference.
Different decision points provide execution histories of
varying lengths.

The reference strategy $s_k^{\mathrm{ref}}$ is the strategy
invoked by the teacher at decision $k$.
Let $y_k$ denote its textual encoding.
The resulting supervised dataset is
$
\mathcal{D}_{\mathrm{coord}}
=
\left\{
(\mathbf{x}_k,y_k)
\right\}.
$ Observations and execution outcomes resulting from
$s_k^{\mathrm{ref}}$ are excluded from $\mathbf{x}_k$.
Likewise, each historical summary is restricted to information available when its corresponding execution ended.

We optimize the reference strategy outputs using
$
\mathcal{L}_{\mathrm{coord}}(\theta)
=
-\mathbb{E}_{\mathcal{D}_{\mathrm{coord}}}
\Bigl[
|y|^{-1}
\sum_{j=1}^{|y|}
\log P_{\theta}(y_j \mid \mathbf{x},y_{<j})
\Bigr],
$
where the expectation is taken over
$(\mathbf{x},y)\sim\mathcal{D}_{\mathrm{coord}}$.

This stage trains the agent to associate execution history
with subsequent strategy invocations.
Dataset statistics are reported in the experimental setup.

\section{Experimental Results}

\subsection{Experimental Setup}

\textbf{Datasets}. We utilize the HM3D\cite{HM3D} and MP3D\cite{MP3D} datasets under the standard Object Goal Navigation (OGN) benchmark. The HM3D validation split consists of 2000 episodes across 20 scenes, encompassing 6 object categories. Similarly, the MP3D validation set contains 2195 episodes spanning 11 scenes and 21 categories. Notably, a statistical analysis of these environments reveals a strong prevalence of multi-story layouts: approximately 65\% of the scenes in HM3D and 73\% in MP3D feature multiple floors.\\

\textbf{Training Datasets.}
For SFT Stage I, we construct a map-centric VQA dataset using
21,029 trajectories from VLN-CE~\cite{krantz2020beyond} and the HM3D and MP3D
environments.
For each trajectory, we sample an average of 20 map
subsequences, each spanning 3--8 consecutive navigation
steps, and construct corresponding question--answer pairs.
This yields 420,780 VQA training samples.
For SFT Stage II, we use GPT-4o as the teacher model to collect
navigation demonstrations in HM3D and MP3D.
We perform approximately 100 rollout attempts for each
navigation segment, obtaining an average of 32.3 valid
successful rollouts and approximately 135,498 valid
trajectories in total.
An automated annotation tool then partitions these
trajectories into strategy-execution segments.
Each trajectory yields an average of 8 supervised
decision samples, resulting in 1,083,984 training samples
for strategy coordination.

\noindent \textbf{Evaluation Metrics}.
To comprehensively evaluate all methods, we employ two standard metrics: Success Rate (SR) and Success weighted by inverse Path Length (SPL). While SR captures the binary outcome of task completion, SPL quantifies the agent's navigational efficiency by comparing its trajectory against the shortest optimal path, assigning a score of zero to unsuccessful episodes.\\

\noindent \textbf{Experimental Details}. We employ RAM\cite{ram} to generate category prompts, while Grounding-DINO\cite{Gd} is utilized for open-vocabulary detection. Object contours are extracted via SAM\cite{sam} segmentation. To facilitate boundary reasoning, we leverage BLIP-2\cite{Blip-2} for semantic similarity representation, Places365\cite{Places} for scene classification, and GPT-4o for LLM-based inference. For all experiments, the system was deployed using two NVIDIA RTX 4090 GPUs, with one GPU dedicated to running the fine-tuned Qwen2.5-VL-7B-Instruct model. We use the AdamW optimizer with a learning rate of
$2\times10^{-5}$ and keep all parameters of the vision
encoder frozen throughout both SFT stages.
Each stage is trained for 1,000 optimizer update steps.
Training across both stages takes approximately 150 hours
on eight NVIDIA A100 GPUs, corresponding to approximately
1,200 GPU-hours.

\subsection{Comparison with SOTA}
As shown in Tab.~\ref{tab:comparison}, we evaluate our proposed LERR-Nav  methods on the HM3D and MP3D datasets. Our method consistently outperforms all existing baselines, establishing a new SOTA performance across both benchmarks in zero-shot settings. 

Specifically, on the HM3D dataset, LERR-Nav achieves an impressive $72.3\%$ in SR and $35.6\%$ in SPL. This represents a significant absolute improvement of $+6.9\%$ in SR and $+2.1\%$ in SPL over the previously best-performing method, ASCENT. Furthermore, on the MP3D dataset, our approach reaches $47.9\%$ SR and $18.7\%$ SPL. Compared to the strong baseline ASCENT, we observe a $+3.4\%$ increase in SR. Additionally, LERR-Nav surpasses BeliefMapNav, the prior best in path efficiency on MP3D, yielding a $+1.1\%$ gain in SPL. 

To validate cross-floor capabilities, we divide the HM3D episodes into intra-floor (The target object and the robot are initialized on the same floor) and inter-floor (The target object and the robot are initialized on the different floor ) scenarios. As shown in Tab. \ref{tab:cross_floor_results}, LERR-Nav's overall enhancement is predominantly driven by its inter-floor breakthrough. Alongside steady intra-floor improvements, our method achieves remarkable cross-level gains, outperforming the strongest baseline, ASCENT, by $+29.2\%$ in SR and $+12.4\%$ in SPL.

\begin{table*}[t]
    \centering
    \caption{Comparison of LERR-Nav and other methods on the HM3D and MP3D datasets.}
    \label{tab:comparison}
    \setlength{\tabcolsep}{7pt}
    \begin{tabular}{l c cc cc}
        \toprule
        \multirow{2}{*}{Method}
        & \multirow{2}{*}{Venue}
        & \multicolumn{2}{c}{HM3D}
        & \multicolumn{2}{c}{MP3D} \\
        \cmidrule(lr){3-4} \cmidrule(lr){5-6}
        & & SR$\uparrow$ & SPL$\uparrow$
          & SR$\uparrow$ & SPL$\uparrow$ \\
        \midrule
        ZSON~\cite{embeddings}
        & NeurIPS'22 & 25.5 & 12.6 & 15.3 & 4.8 \\
        ESC~\cite{Esc}
        & ICML'23 & 39.2 & 22.3 & 28.7 & 14.2 \\
        L3MVN~\cite{L3mvn}
        & IROS'23 & 50.4 & 23.1 & 34.9 & 14.5 \\
        PSL~\cite{PSL}
        & ECCV'24 & 42.4 & 19.2 & 18.9 & 6.9 \\
        VLFM~\cite{Vlfm}
        & ICRA'24 & 52.5 & 30.4 & 36.4 & 17.5 \\
        OpenFMNav~\cite{Openfmnav}
        & NAACL'24 & 52.5 & 24.1 & 37.2 & 15.7 \\
        SG-NAV~\cite{Sg-nav}
        & NeurIPS'24 & 54.0 & 24.9 & 40.2 & 16.0 \\
        BeliefMapNav~\cite{Beliefmapnav}
        & NeurIPS'25 & 61.4 & 30.6 & 37.3 & 17.6 \\
        ASCENT~\cite{Stairway}
        & RAL'26 & 65.4 & 33.5 & 44.5 & 15.5 \\
        \midrule
        \textbf{LERR-Nav}
        & \textbf{ours}
        & \textbf{72.3} & \textbf{35.6}
        & \textbf{47.9} & \textbf{18.7} \\
        \bottomrule
    \end{tabular}
\end{table*}

\begin{table}[htbp]
\centering
\caption{Performance comparison between intra and inter-floor navigation scenarios.}
\label{tab:cross_floor_results}
\setlength{\tabcolsep}{5pt}
\begin{tabular}{l | cc | cc | cc}
\toprule
\multirow{2}{*}{\textbf{Method}} & \multicolumn{2}{c|}{\textbf{Intra-Floor Scenarios}} & \multicolumn{2}{c|}{\textbf{Inter-Floor Scenarios}} & \multicolumn{2}{c}{\textbf{All Episodes}} \\
 & \textbf{SR}$\uparrow$ & \textbf{SPL}$\uparrow$ & \textbf{SR}$\uparrow$ & \textbf{SPL}$\uparrow$ & \textbf{SR}$\uparrow$ & \textbf{SPL}$\uparrow$ \\
\midrule
VLFM\cite{Vlfm} & 64.6 & 37.3 & 0.4 & 0.1 & 52.8 & 30.5 \\
MFNP\cite{MFNP} & 68.4 & 30.5 & 13.4 & 9.8 & 58.3 & 26.7 \\
ASCENT\cite{Stairway} & 72.6 & 37.7 & 33.3 & 14.9 & 65.4 & 33.5 \\
\textbf{LERR-Nav} & \textbf{75.8} & \textbf{38.1} & \textbf{62.5} & \textbf{27.3} & \textbf{72.3} & \textbf{35.6} \\ 
\bottomrule
\end{tabular}
\end{table}

\subsection{Ablative Study}
To comprehensively validate the efficacy of the proposed components in LERR-Nav, we conduct extensive ablation study. Additional ablation studies are provided in the Appendix~\ref{sec:Experimental Ablative Study Details}.

\paragraph{Impact of Our High-Level Policy.}
We evaluate different variants of the high-level policy on HM3D
in Table~\ref{tab:high_level_policy}.
Stage-I map-centric SFT improves the base model from
43.3/19.6 to 58.1/27.3 in SR/SPL, outperforming GPT-5.
Stage-II coordination SFT further increases performance to
72.3/35.6.
Overall, the two-stage training yields gains of 29 SR and
16 SPL points over the base model, validating the benefits
of both map-centric grounding and strategy-coordination training.

\begin{table}[t]
    \centering
    \caption{Impact of the high-level coordination policy on HM3D.}
    \label{tab:high_level_policy}
    \setlength{\tabcolsep}{7pt}
    \begin{tabular}{lcc}
        \toprule
        High-Level Policy & SR$\uparrow$ & SPL$\uparrow$ \\
        \midrule
        GPT-5                         & 52.2 & 23.2 \\
        LERR-Nav (Base)               & 43.3 & 19.6 \\
        LERR-Nav (Stage-I SFT)        & 58.1 & 27.3 \\
        \textbf{LERR-Nav (Stage-I+II SFT)}
                                      & \textbf{72.3} & \textbf{35.6} \\
        \bottomrule
    \end{tabular}
\end{table}

\paragraph{Impact of Navigation Strategies.}
Table~\ref{tab:strategy_ablation} shows that removing any
strategy degrades navigation performance.
LLM-guided exploration contributes the most, as its removal
reduces SR and SPL by 20.5 and 5.4 points, respectively.
Removing value-function-guided exploration decreases SR by
8.8 points, while removing recovery causes a 7.1-point SR
drop and a 3.7-point SPL drop.
Reminiscing also provides consistent gains, improving SR and
SPL by 6.0 and 3.1 points.
These results demonstrate that the four strategies are
complementary, with the complete model achieving the best
performance of 72.3 SR and 35.6 SPL.

\begin{table}[t]
    \centering
    \caption{Ablation of the navigation strategies on HM3D.}
    \label{tab:strategy_ablation}
    \setlength{\tabcolsep}{4pt}
    \begin{tabular}{lcccccc}
        \toprule
        Variant & VF-E & LLM-E & Rec. & Rem. &
        SR$\uparrow$ & SPL$\uparrow$ \\
        \midrule
        w/o VF-E   & -- & \checkmark & \checkmark & \checkmark & 63.5 & 32.6 \\
        w/o LLM-E  & \checkmark & -- & \checkmark & \checkmark & 51.8 & 30.2 \\
        w/o Rec.   & \checkmark & \checkmark & -- & \checkmark & 65.2 & 31.9 \\
        w/o Rem.   & \checkmark & \checkmark & \checkmark & -- & 66.3 & 32.5 \\
        \midrule
        \textbf{Full}
        & \checkmark & \checkmark & \checkmark & \checkmark
        & \textbf{72.3} & \textbf{35.6} \\
        \bottomrule
    \end{tabular}
\end{table}

\subsection{Qualitative Analysis}
\begin{figure*}[h]
    \centering
    \includegraphics[width=\textwidth]{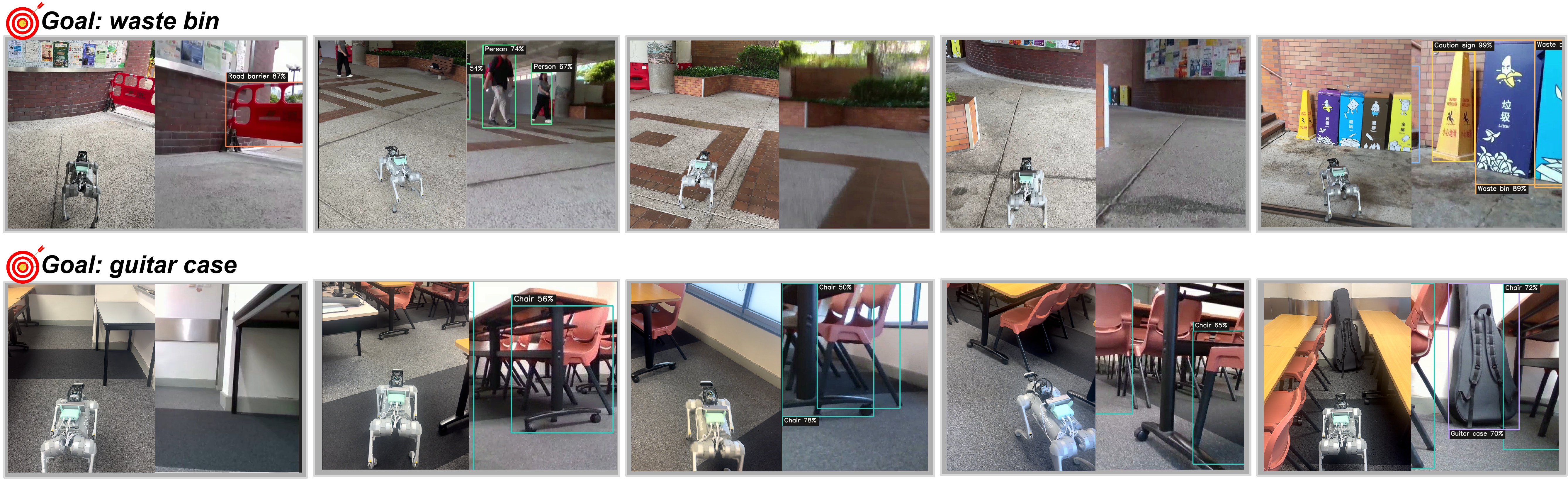}
    \caption{
    Real-world deployment of LERR-Nav on a Unitree Go2
    quadruped robot.
    }
    \label{fig:qualitative_real}
\end{figure*}

To further demonstrate the effectiveness of our LERR-Nav, we compare it against a strong recent baseline  ASCENT\cite{Stairway} on the same navigation episode (See in Appendix~\ref{sec:Experimental Demo Details}). 
Real-world deployment on the Unitree Go2
(Fig.~\ref{fig:qualitative_real}) further demonstrates the effectiveness of our method in sim-to-real transfer.

\section{Conclusion}
We presented LERR-Nav, which learns to coordinate four
complementary navigation strategies using execution memory
and a high-level vision-language policy.
A two-stage SFT curriculum progressively develops spatial
understanding and strategy coordination.
Experiments on HM3D and MP3D validate its effectiveness,
while deployment on a Unitree Go2 demonstrates its potential
for sim-to-real object navigation.

\section*{AI use strategyment}
In this work, we used generative AI tools (e.g., large language models) for manuscript language editing, grammar refinement, and code debugging. We have not used generative AI tools for generating core research ideas, designing methodologies, or creating experimental data. We have thoroughly reviewed all AI-assisted text and code to ensure their accuracy, correctness, and originality. We take full responsibility for the final content of this work, including all text, claims, and artifacts.

\bibliography{iclr2027_conference}

@String(CVPR  = {IEEE Conf. Comput. Vis. Pattern Recog.})

@String(ICCV  = {Int. Conf. Comput. Vis.})

@String(ECCV  = {Eur. Conf. Comput. Vis.})

@String(NeurIPS = {Adv. Neural Inform. Process. Syst.})

@String(ICML  = {Int. Conf. Mach. Learn.})

@String(ICPR  = {Int. Conf. Pattern Recog.})

@String(CVPR  = {CVPR})

@String(ICCV  = {ICCV})

@String(ECCV  = {ECCV})

@String(NeurIPS = {NeurIPS})

@String(ICML  = {ICML})

@String(ICPR  = {ICPR})

@article{SSR-ZS,
  title={SSR-ZSON: Zero-Shot Object Navigation via Spatial-Semantic Relations within a Hierarchical Exploration Framework},
  author={Meng, Xiangyi and Li, Delun and Mao, Zihao and Yang, Yi and Song, Wenjie},
  journal={arXiv preprint \href{https://arxiv.org/abs/2509.24763}{arXiv:2509.24763}},
  year={2025}
}

@article{Sg-nav,
  title={Sg-nav: Online 3d scene graph prompting for llm-based zero-shot object navigation},
  author={Yin, Hang and Xu, Xiuwei and Wu, Zhenyu and Zhou, Jie and Lu, Jiwen},
  journal={Advances in Neural Information Processing Systems (NeurIPS)},
  volume={37},
  pages={5285--5307},
  year={2024}
}

@article{Apexnav,
  title={Apexnav: An adaptive exploration strategy for zero-shot object navigation with target-centric semantic fusion},
  author={Zhang, Mingjie and Du, Yuheng and Wu, Chengkai and Zhou, Jinni and Qi, Zhenchao and Ma, Jun and Zhou, Boyu},
  journal={IEEE Robotics and Automation Letters},
  year={2025},
  publisher={IEEE}
}

@article{Stairway,
  title={Stairway to Success: An Online Floor-Aware Zero-Shot Object-Goal Navigation Framework via LLM-Driven Coarse-to-Fine Exploration},
  author={Gong, Zeying and Li, Rong and Hu, Tianshuai and Qiu, Ronghe and Kong, Lingdong and Zhang, Lingfeng and Zhao, Guoyang and Ding, Yiyi and Liang, Junwei},
  journal={IEEE Robotics and Automation Letters},
  year={2026},
  publisher={IEEE}
}

@inproceedings{Vlfm,
  title={Vlfm: Vision-language frontier maps for zero-shot semantic navigation},
  author={Yokoyama, Naoki and Ha, Sehoon and Batra, Dhruv and Wang, Jiuguang and Bucher, Bernadette},
  booktitle={2024 IEEE International Conference on Robotics and Automation (ICRA)},
  pages={42--48},
  year={2024},
  Xorganization={IEEE}
}

@inproceedings{3D-mem,
  title={3D-mem: 3D scene memory for embodied exploration and reasoning},
  author={Yang, Yuncong and Yang, Han and Zhou, Jiachen and Chen, Peihao and Zhang, Hongxin and Du, Yilun and Gan, Chuang},
  booktitle={Proceedings of the IEEE/CVF Conference on Computer Vision and Pattern Recognition (CVPR)},
  pages={17294--17303},
  year={2025}
}

@inproceedings{train1,
  title={Poni: Potential functions for objectgoal navigation with interaction-free learning},
  author={Ramakrishnan, Santhosh Kumar and Chaplot, Devendra Singh and Al-Halah, Ziad and Malik, Jitendra and Grauman, Kristen},
  booktitle={Proceedings of the IEEE/CVF Conference on Computer Vision and Pattern Recognition (CVPR)},
  pages={18890--18900},
  year={2022}
}

@inproceedings{train2,
  title={Pirlnav: Pretraining with imitation and rl finetuning for objectnav},
  author={Ramrakhya, Ram and Batra, Dhruv and Wijmans, Erik and Das, Abhishek},
  booktitle={Proceedings of the IEEE/CVF Conference on Computer Vision and Pattern Recognition (CVPR)},
  pages={17896--17906},
  year={2023}
}

@inproceedings{train3,
  title={Object memory transformer for object goal navigation},
  author={Fukushima, Rui and Ota, Kei and Kanezaki, Asako and Sasaki, Yoko and Yoshiyasu, Yusuke},
  booktitle={2022 International Conference on Robotics and Automation (ICRA)},
  pages={11288--11294},
  year={2022},
  organization={IEEE}
}

@inproceedings{train4,
  title={Thda: Treasure hunt data augmentation for semantic navigation},
  author={Maksymets, Oleksandr and Cartillier, Vincent and Gokaslan, Aaron and Wijmans, Erik and Galuba, Wojciech and Lee, Stefan and Batra, Dhruv},
  booktitle={Proceedings of the IEEE/CVF International Conference on Computer Vision (ICCV)},
  pages={15374--15383},
  year={2021}
}

@article{train5,
  title={Efficient robotic object search via hiem: Hierarchical policy learning with intrinsic-extrinsic modeling},
  author={Ye, Xin and Yang, Yezhou},
  journal={IEEE Robotics and Automation Letters},
  volume={6},
  number={3},
  pages={4425--4432},
  year={2021},
  publisher={IEEE}
}

@inproceedings{Openfmnav,
  title={Openfmnav: Towards open-set zero-shot object navigation via vision-language foundation models},
  author={Kuang, Yuxuan and Lin, Hai and Jiang, Meng},
  booktitle={Findings of the Association for Computational Linguistics: NAACL 2024},
  pages={338--351},
  year={2024}
}

@article{Cl-cotnav,
  title={Cl-cotnav: Closed-loop hierarchical chain-of-thought for zero-shot object-goal navigation with vision-language models},
  author={Cai, Yuxin and He, Xiangkun and Wang, Maonan and Guo, Hongliang and Yau, Wei-Yun and Lv, Chen},
  journal={arXiv preprint \href{https://arxiv.org/abs/2504.09000}{arXiv:2504.09000}},
  year={2025}
}

@article{SemNav,
  title={SemNav: A Model-Based Planner for Zero-Shot Object Goal Navigation Using Vision-Foundation Models},
  author={Debnath, Arnab and Stein, Gregory J and Kosecka, Jana},
  journal={arXiv preprint \href{https://arxiv.org/abs/2506.03516}{arXiv:2506.03516}},
  year={2025}
}

@article{Vln-game,
  title={Vln-game: Vision-language equilibrium search for zero-shot semantic navigation},
  author={Yu, Bangguo and Liu, Yuzhen and Han, Lei and Kasaei, Hamidreza and Li, Tingguang and Cao, Ming},
  journal={arXiv preprint \href{https://arxiv.org/abs/2411.11609}{arXiv:2411.11609}},
  year={2024}
}

@article{Loc-zson,
  title={Loc-zson: Language-driven object-centric zero-shot object retrieval and navigation},
  author={Guan, Tianrui and Yang, Yurou and Cheng, Harry and Lin, Muyuan and Kim, Richard and Madhivanan, Rajasimman and Sen, Arnie and Manocha, Dinesh},
  journal={arXiv preprint \href{https://arxiv.org/abs/2405.05363}{arXiv:2405.05363}},
  year={2024}
}

@inproceedings{L-ZSON,
  title={Zero-shot object navigation with vision-language models reasoning},
  author={Wen, Congcong and Huang, Yisiyuan and Huang, Hao and Huang, Yanjia and Yuan, Shuaihang and Hao, Yu and Lin, Hui and Liu, Yu-Shen and Fang, Yi},
  booktitle={International Conference on Pattern Recognition (ICPR)},
  pages={389--404},
  year={2025},
  organization={Springer}
}

@inproceedings{Blip-2,
  title={Blip-2: Bootstrapping language-image pre-training with frozen image encoders and large language models},
  author={Li, Junnan and Li, Dongxu and Savarese, Silvio and Hoi, Steven},
  booktitle={International conference on machine learning},
  pages={19730--19742},
  year={2023},
  organization={PMLR}
}

@article{HM3D,
  title={Habitat-matterport 3d dataset (hm3d): 1000 large-scale 3d environments for embodied ai},
  author={Ramakrishnan, Santhosh K and Gokaslan, Aaron and Wijmans, Erik and Maksymets, Oleksandr and Clegg, Alex and Turner, John and Undersander, Eric and Galuba, Wojciech and Westbury, Andrew and Chang, Angel X and others},
  journal={arXiv preprint \href{https://arxiv.org/abs/2109.08238}{arXiv:2109.08238}},
  year={2021}
}

@article{MP3D,
  title={Matterport3d: Learning from rgb-d data in indoor environments},
  author={Chang, Angel and Dai, Angela and Funkhouser, Thomas and Halber, Maciej and Niessner, Matthias and Savva, Manolis and Song, Shuran and Zeng, Andy and Zhang, Yinda},
  journal={arXiv preprint \href{https://arxiv.org/abs/1709.06158}{arXiv:1709.06158}},
  year={2017}
}

@article{Obj-task,
  title={Objectnav revisited: On evaluation of embodied agents navigating to objects},
  author={Batra, Dhruv and Gokaslan, Aaron and Kembhavi, Aniruddha and Maksymets, Oleksandr and Mottaghi, Roozbeh and Savva, Manolis and Toshev, Alexander and Wijmans, Erik},
  journal={arXiv preprint \href{https://arxiv.org/abs/2006.13171}{arXiv:2006.13171}},
  year={2020}
}

@article{SpatialNav,
  title={SpatialNav: Leveraging Spatial Scene Graphs for Zero-Shot Vision-and-Language Navigation},
  author={Zhang, Jiwen and Li, Zejun and Wang, Siyuan and Shi, Xiangyu and Wei, Zhongyu and Wu, Qi},
  journal={arXiv preprint \href{https://arxiv.org/abs/2601.06806}{arXiv:2601.06806}},
  year={2026}
}

@article{3DGSNav,
  title={3DGSNav: Enhancing Vision-Language Model Reasoning for Object Navigation via Active 3D Gaussian Splatting},
  author={Zheng, Wancai and Chen, Hao and Lu, Xianlong and Ou, Linlin and Yu, Xinyi},
  journal={arXiv preprint \href{https://arxiv.org/abs/2602.12159}{arXiv:2602.12159}},
  year={2026}
}

@article{Open_scene,
  title={Open scene graphs for open-world object-goal navigation},
  author={Loo, Joel and Wu, Zhanxin and Hsu, David},
  journal={The International Journal of Robotics Research},
  Xpages={02783649251369549},
  year={2025},
  url={https://doi.org/10.1177/02783649251369549},
  Xpublisher={SAGE Publications Sage UK: London, England}
}

@article{exp_pro1,
  title={Handle object navigation as weighted traveling repairman problem},
  author={Liu, Ruimeng and Xu, Xinhang and Yuan, Shenghai and Xie, Lihua},
  journal={arXiv preprint \href{https://arxiv.org/abs/2503.06937}{arXiv:2503.06937}},
  year={2025}
}

@article{Beliefmapnav,
  title={Beliefmapnav: 3d voxel-based belief map for zero-shot object navigation},
  author={Zhou, Zibo and Hu, Yue and Zhang, Lingkai and Li, Zonglin and Chen, Siheng},
  journal={arXiv preprint \href{https://arxiv.org/abs/2506.06487}{arXiv:2506.06487}},
  year={2025}
}

@article{Curiosity,
  title={Curiosity-Driven Zero-Shot Object Navigation With Vision-Language Models},
  author={Gu, Zhengyi and Zhou, Yuheng and Xue, Yuquan and Wang, Hanzhang},
  journal={IEEE Robotics and Automation Letters},
  year={2026},
  publisher={IEEE}
}

@inproceedings{Esc,
  title={Esc: Exploration with soft commonsense constraints for zero-shot object navigation},
  author={Zhou, Kaiwen and Zheng, Kaizhi and Pryor, Connor and Shen, Yilin and Jin, Hongxia and Getoor, Lise and Wang, Xin Eric},
  booktitle={International Conference on Machine Learning (ICML)},
  pages={42829--42842},
  year={2023},
  organization={PMLR}
}

@article{Instructnav,
  title={Instructnav: Zero-shot system for generic instruction navigation in unexplored environment},
  author={Long, Yuxing and Cai, Wenzhe and Wang, Hongcheng and Zhan, Guanqi and Dong, Hao},
  journal={arXiv preprint \href{https://arxiv.org/abs/2406.04882}{arXiv:2406.04882}},
  year={2024}
}

@article{Imaginenav,
  title={Imaginenav: Prompting vision-language models as embodied navigator through scene imagination},
  author={Zhao, Xinxin and Cai, Wenzhe and Tang, Likun and Wang, Teng},
  journal={arXiv preprint \href{https://arxiv.org/abs/2410.09874}{arXiv:2410.09874}},
  year={2024}
}

@inproceedings{HuLE-Nav,
  title={HuLE-Nav: Human-Like Exploration for Zero-Shot Object Navigation via Vision-Language Models},
  author={Han, Peilong and Zhang, Min and Tang, Hongyao and ZHENG, YAN and others},
  booktitle={NeurIPS 2024 Workshop on Behavioral Machine Learning},
  Xbooktitle={NeurIPS Workshops},
  year={2024}
}

@inproceedings{Cows,
  title={Cows on pasture: Baselines and benchmarks for language-driven zero-shot object navigation},
  author={Gadre, Samir Yitzhak and Wortsman, Mitchell and Ilharco, Gabriel and Schmidt, Ludwig and Song, Shuran},
  booktitle={Proceedings of the IEEE/CVF Conference on Computer Vision and Pattern Recognition (CVPR)},
  pages={23171--23181},
  year={2023}
}

@article{Orb-slam2,
  title={Orb-slam2: An open-source slam system for monocular, stereo, and rgb-d cameras},
  author={Mur-Artal, Raul and Tard{\'o}s, Juan D},
  journal={IEEE Transactions on Robotics (TRO)},
  volume={33},
  number={5},
  pages={1255--1262},
  year={2017},
  publisher={IEEE}
}

@inproceedings{ram,
  title={Recognize anything: A strong image tagging model},
  author={Zhang, Youcai and Huang, Xinyu and Ma, Jinyu and Li, Zhaoyang and Luo, Zhaochuan and Xie, Yanchun and Qin, Yuzhuo and Luo, Tong and Li, Yaqian and Liu, Shilong and others},
  booktitle={Proceedings of the IEEE/CVF Conference on Computer Vision and Pattern Recognition (CVPR)},
  pages={1724--1732},
  year={2024}
}

@article{Places,
  title={Places: A 10 million image database for scene recognition},
  author={Zhou, Bolei and Lapedriza, Agata and Khosla, Aditya and Oliva, Aude and Torralba, Antonio},
  journal={IEEE Transactions on Pattern Analysis and Machine Intelligence (TPAMI)},
  volume={40},
  number={6},
  pages={1452--1464},
  year={2017},
  publisher={IEEE}
}

@inproceedings{Gd,
  title={Grounding dino: Marrying dino with grounded pre-training for open-set object detection},
  author={Liu, Shilong and Zeng, Zhaoyang and Ren, Tianhe and Li, Feng and Zhang, Hao and Yang, Jie and Jiang, Qing and Li, Chunyuan and Yang, Jianwei and Su, Hang and others},
  booktitle={European Conference on Computer Vision (ECCV)},
  pages={38--55},
  year={2024},
  organization={Springer}
}

@inproceedings{sam,
  title={Segment anything},
  author={Kirillov, Alexander and Mintun, Eric and Ravi, Nikhila and Mao, Hanzi and Rolland, Chloe and Gustafson, Laura and Xiao, Tete and Whitehead, Spencer and Berg, Alexander C and Lo, Wan-Yen and others},
  booktitle={Proceedings of the IEEE/CVF International Conference on Computer Vision (ICCV)},
  pages={4015--4026},
  year={2023}
}

@inproceedings{L3mvn,
  title={L3mvn: Leveraging large language models for visual target navigation},
  author={Yu, Bangguo and Kasaei, Hamidreza and Cao, Ming},
  booktitle={2023 IEEE/RSJ International Conference on Intelligent Robots and Systems (IROS)},
  pages={3554--3560},
  year={2023},
  organization={IEEE}
}

@inproceedings{PSL,
  title={Prioritized semantic learning for zero-shot instance navigation},
  author={Sun, Xinyu and Liu, Lizhao and Zhi, Hongyan and Qiu, Ronghe and Liang, Junwei},
  booktitle={European Conference on Computer Vision (ECCV)},
  pages={161--178},
  year={2024},
  organization={Springer}
}

@inproceedings{MFNP,
  title={Multi-floor zero-shot object navigation policy},
  author={Zhang, Lingfeng and Wang, Hao and Xiao, Erjia and Zhang, Xinyao and Zhang, Qiang and Jiang, Zixuan and Xu, Renjing},
  booktitle={2025 IEEE International Conference on Robotics and Automation (ICRA)},
  pages={6416--6422},
  year={2025},
  organization={IEEE}
}

@article{embeddings,
  title={Zson: Zero-shot object-goal navigation using multimodal goal embeddings},
  author={Majumdar, Arjun and Aggarwal, Gunjan and Devnani, Bhavika and Hoffman, Judy and Batra, Dhruv},
  journal={Advances in Neural Information Processing Systems (NeurIPS)},
  volume={35},
  pages={32340--32352},
  year={2022}
}

@inproceedings{Clip,
  title={Learning transferable visual models from natural language supervision},
  author={Radford, Alec and Kim, Jong Wook and Hallacy, Chris and Ramesh, Aditya and Goh, Gabriel and Agarwal, Sandhini and Sastry, Girish and Askell, Amanda and Mishkin, Pamela and Clark, Jack and others},
  booktitle={International Conference on Machine Learning (ICML)},
  pages={8748--8763},
  year={2021},
  organization={PMLR}
}

@inproceedings{blip,
  title={Blip: Bootstrapping language-image pre-training for unified vision-language understanding and generation},
  author={Li, Junnan and Li, Dongxu and Xiong, Caiming and Hoi, Steven},
  booktitle={International conference on machine learning},
  pages={12888--12900},
  year={2022},
  organization={PMLR}
}

@inproceedings{Cognav,
  title={Cognav: Cognitive process modeling for object goal navigation with llms},
  author={Cao, Yihan and Zhang, Jiazhao and Yu, Zhinan and Liu, Shuzhen and Qin, Zheng and Zou, Qin and Du, Bo and Xu, Kai},
  booktitle={Proceedings of the IEEE/CVF International Conference on Computer Vision (ICCV)},
  pages={9550--9560},
  year={2025}
}

@inproceedings{SemExp,
  title={Object Goal Navigation using Goal-Oriented Semantic Exploration},
  author={Chaplot, Devendra Singh and Gandhi, Dhiraj and
            Gupta, Abhinav and Salakhutdinov, Ruslan},
  booktitle={Advances in Neural Information Processing Systems (NeurIPS)},
  Xbooktitle={NeurIPS},
  year={2020}
}

@article{RIM,
  title={Object Goal Navigation with Recursive Implicit Maps},
  author={Shizhe Chen and Thomas Chabal and Ivan Laptev and Cordelia Schmid},
  journal={2023 IEEE/RSJ International Conference on Intelligent Robots and Systems (IROS)},
  year={2023},
  pages={7089--7096},
  organization={IEEE},
  Xurl={https://api.semanticscholar.org/CorpusID:260775848}
}

@inproceedings{XGX,
    author = {{Wasserman}, Justin and {Chowdhary}, Girish and {Gupta}, Abhinav and {Jain}, Unnat},
    title = "{Exploitation-Guided Exploration for Semantic Embodied Navigation}",
    booktitle = {2024 IEEE International Conference on Robotics and Automation (ICRA)},
    year = {2024},
    pages = {2901--2908},
}

@ARTICLE{OVG-Nav,
    author={Yoo, Hwiyeon and Choi, Yunho and Park, Jeongho and Oh, Songhwai},
    journal={IEEE Robotics and Automation Letters}, 
    title={Commonsense-Aware Object Value Graph for Object Goal Navigation}, 
    year={2024},
    volume={9},
    number={5},
    pages={4423-4430},
}

@inproceedings{SEEK,
    title = {SEEK: Semantic Reasoning for Object Goal Navigation in Real World Inspection Tasks},
    author = {Muhammad Fadhil Ginting and Sung-Kyun Kim and David D. Fan and Matteo Palieri and Mykel J. Kochenderfer and Ali-akbar Agha-mohammadi},
    year = {2024},
    booktitle = {Proc. of Robotics: Science and Systems (RSS)}
}

@article{RL-OGN,
    title={What Matters in RL-Based Methods for Object-Goal Navigation? An Empirical Study and A Unified Framework}, 
    author={Hongze Wang and Boyang Sun and Jiaxu Xing and Fan Yang and Marco Hutter and Dhruv Shah and Davide Scaramuzza and Marc Pollefeys},
    year={2025},
    journal={arXiv preprint \href{https://arxiv.org/abs/2510.01830}{arXiv:2510.01830}},
}

@article{bai2025qwen25vl,
  title   = {{Qwen2.5-VL} Technical Report},
  author  = {Bai, Shuai and
             Chen, Keqin and
             Liu, Xuejing and
             Wang, Jialin and
             Ge, Wenbin and
             Song, Sibo and
             Dang, Kai and
             Wang, Peng and
             Wang, Shijie and
             Tang, Jun and
             Zhong, Humen and
             Zhu, Yuanzhi and
             Yang, Mingkun and
             Li, Zhaohai and
             Wan, Jianqiang and
             Wang, Pengfei and
             Ding, Wei and
             Fu, Zheren and
             Xu, Yiheng and
             Ye, Jiabo and
             Zhang, Xi and
             Xie, Tianbao and
             Cheng, Zesen and
             Zhang, Hang and
             Yang, Zhibo and
             Xu, Haiyang and
             Lin, Junyang},
  journal = {arXiv preprint arXiv:2502.13923},
  year    = {2025},
  doi     = {10.48550/arXiv.2502.13923},
  url     = {https://arxiv.org/abs/2502.13923}
}

@misc{openai2026gpt6astra,
  author       = {{OpenAI}},
  title        = {{GPT-6 Astra Model}},
  year         = {2026},
  howpublished = {\url{https://developers.openai.com/api/docs/models/gpt-6-astra}},
  note         = {OpenAI API documentation. Accessed: September 26, 2026}
}

@inproceedings{wijmans2022ver,
  title     = {{VER}: Scaling On-Policy {RL} Leads to the Emergence
               of Navigation in Embodied Rearrangement},
  author    = {Wijmans, Erik and Essa, Irfan A. and Batra, Dhruv},
  booktitle = {Advances in Neural Information Processing Systems},
  volume    = {35},
  pages     = {7727--7740},
  year      = {2022}
}

@inproceedings{krantz2020beyond,
  title     = {Beyond the Nav-Graph: Vision-and-Language Navigation
               in Continuous Environments},
  author    = {Krantz, Jacob and Wijmans, Erik and Majumdar, Arjun
               and Batra, Dhruv and Lee, Stefan},
  booktitle = {European Conference on Computer Vision (ECCV)},
  year      = {2020}
}
\bibliographystyle{iclr2027_conference}

\newpage
\appendix
\section{Appendix}
\label{sec:TD}

\subsection{Exploration Value Map Details}
 We define a value  function $V(f)$ for each frontier $f \in \mathcal{F_\text{floor}}$ as follows:
\begin{equation}
\label{eq1}
V(f) = \alpha \cdot S_\text{sem}(f) + \beta \cdot S_\text{dist}(f),
\end{equation}

The semantic relevance score of a frontier $f$, denoted as $S_{\text{sem}}(f)$, can be computed using BLIP-2~\cite{Blip-2} by measuring the cosine similarity between the RGB observation $I_f$ at the frontier and a set of context-aware text prompts $T$ (e.g., \text{bathtub} or \text{sink} when the target is \text{toilet}) generated by the LLM's common-sense reasoning:
\begin{equation}
S_{\text{sem}}(f) = \max_{t \in T} \, \text{cos\_sim} \big( \phi(I_f), \psi(t) \big),
\end{equation}
where $\phi(\cdot)$ and $\psi(\cdot)$ denote the visual and textual encoders of BLIP-2, respectively, and $\text{cos\_sim}(\mathbf{a}, \mathbf{b}) = \frac{\mathbf{a} \cdot \mathbf{b}}{\|\mathbf{a}\| \, \|\mathbf{b}\|}$ is the cosine similarity between feature vectors.

\subsection{Value-Function-Guided Exploration Strategy  Details}
\begin{figure}[h]
    \centering
    \includegraphics[width=\textwidth]{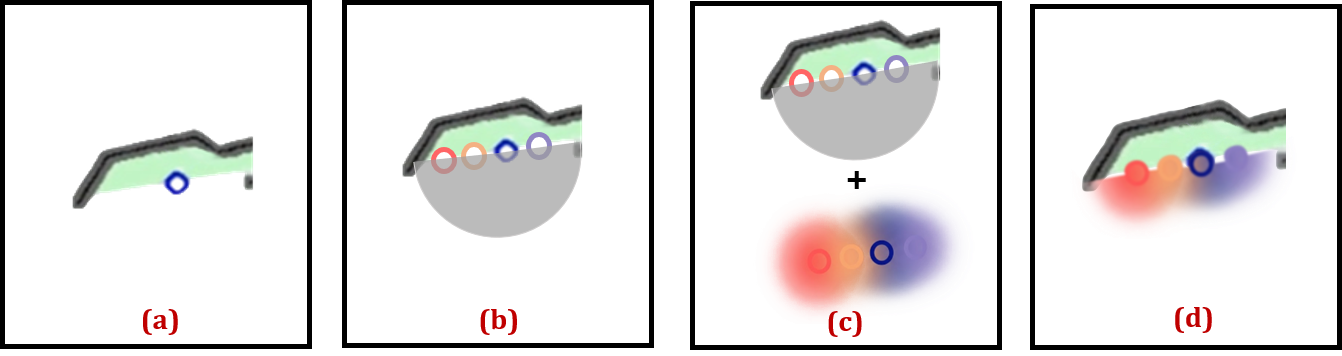}
    \caption{The uncertainty computation process. (a) Initial frontier point at the exploration boundary. (b) FOV projection defining the observable area $S_1$ and uniformly sampled boundary points. (c) Fusion process with the Gaussian kernel function and the assigned value scores for each sampled boundary point. (d) Final Estimated Uncertainty, restricted to the region within $S_1$.}
    \label{fig:7}
\end{figure}
\label{sec:Value-Function-Guided Exploration Strategy  Details}
The computation of uncertainty is illustrated in a visual manner. As shown in Fig.~\ref{fig:7}(a), a frontier point emerges at the boundary between explored and unexplored regions. Subsequently, as illustrated in Fig.~\ref{fig:7}(b), a field-of-view (FOV) rotation projection is applied to the frontier point, resulting in the maximum observable area \(S_1\) associated with this point. For the boundary region \(L\), additional boundary points are sampled uniformly. As depicted in Fig.~\ref{fig:7}(c), each boundary point is assigned a value score, represented with different colors, then, a Gaussian kernel function is applied to each boundary point. The scores are accumulated for points falling within the region \(S_1\), while the scores for points outside \(S_1\) are set to zero. Fig.~\ref{fig:7}(d) demonstrates the final result. It should also be noted that, the Gaussian kernel function is defined as:

\begin{equation}
G(x) = \sum_{i} v_i \exp\Bigg(-\frac{\|x - x_i\|^2}{2\sigma^2}\Bigg),
\end{equation}
where \(x_i\) denotes the location of the \(i\)-th boundary point (serving as the center of the Gaussian), \(v_i\) is the value score of that boundary point, \(x\) is a location within the region \(S_1\), and \(\sigma\) controls the spread of the Gaussian. 

Finally, the pseudo-code of Fast-Thinking is shown in Algorithm \ref{alg:fast_thinking}.

\begin{algorithm}[h]
\caption{Fast-Thinking Mode for Frontier Selection}
\label{alg:fast_thinking}
\begin{algorithmic}[1]
\Require 
    Observations $\mathcal{Z}_k$ up to step $k$, 
    Set of candidate floor frontiers $\mathcal{F}_{\text{floor}}$, 
    Total environment size $|E|$, 
    Max steps $K_{\text{max}}$, 
    Total frontiers $N_{\text{total}}$,
    Gaussian spread parameter $\sigma$
\Ensure Optimal frontier $f^*$ for the next step

\State \textbf{Compute Exploration Reward} $\text{ER}(k)$:
    \State \quad $\text{ER}(k) \gets \sigma_1 \cdot \frac{|U(k)|}{|E|} + \sigma_2 \cdot \frac{N_{\text{frontier}}(k)}{N_{\text{total}}} + \sigma_3 \cdot \left(1 - \frac{k}{K_{\text{max}}}\right)$
\State \textbf{Update balancing factors}:
    \State \quad $\beta(k) \gets \beta_{\text{max}} \cdot \text{ER}(k)$
    \State \quad $\alpha(k) \gets \alpha_{\text{min}} \cdot (1 - \text{ER}(k))$
\State Initialize maximum objective value $J_{\text{max}} \gets -\infty$ and $f^* \gets \emptyset$

\For{each candidate frontier $f \in \mathcal{F}_{\text{floor}}$}
    \State Obtain prior value $V(f)$
    \State Apply FOV rotation projection to $f$ to obtain maximum observable area $S_1(f)$
    \State Extract the boundary region $L$ of $S_1(f)$
    \State Sample boundary points $\{x_i\}$ uniformly from $L$
    \State Assign a value score $v_i$ to each sampled boundary point $x_i$
    
    \State \textbf{Estimate spatial information density} $d(\mathbf{p})$ for points $\mathbf{p} \in S_1(f)$:
        \State \quad $d(\mathbf{p}) \gets \sum_{i} v_i \exp\left(-\frac{\|\mathbf{p} - x_i\|^2}{2\sigma^2}\right)$
    \State \quad \textit{Note: $d(\mathbf{p}) \gets 0$ for points $\mathbf{p} \notin S_1(f)$}
    
    \State \textbf{Compute expected uncertainty reduction}:
        \State \quad $\mathcal{I}(f | \mathcal{F}) \gets \int_{S_1(f)} d(\mathbf{p}) d\mathbf{p} + \sum_{f_j \in \mathcal{F}, j \neq f} \text{Area}\left( S_1(f) \cap S_1(f_j) \right)$
    \State \textbf{Evaluate objective function}:
        \State \quad $J(f) \gets \alpha(k) \cdot V(f) + \beta(k) \cdot \mathcal{I}(f | \mathcal{F})$
    
    \If{$J(f) > J_{\text{max}}$}
        \State $J_{\text{max}} \gets J(f)$
        \State $f^* \gets f$
    \EndIf
\EndFor

\State \Return $f^*$
\end{algorithmic}
\end{algorithm}

\subsection{LLM Prompts}
\label{sec:MP}
In this section, we will provide detailed LLM prompts. These prompts are specifically designed around the robot's three strategys: Exploration strategy, Recovery strategy, and Reminiscing strategy.

\subsubsection{Exploration strategy Prompts}

\begin{tcolorbox}[
    coltitle=white,         
    title=\large{\scriptsize Exploration strategy Prompts}, 
    arc=0mm,                
    boxrule=0mm,            
    toptitle=1mm,           
    bottomtitle=1mm,        
    left=4mm,               
    right=4mm,              
    top=2mm,                
    bottom=2mm,             
    breakable,
    fontupper=\itshape\scriptsize     
]  
    Task: You are a Exploration Planner Assistant. Analyze the image and select the most likely waypoint where the target object (e.g., TV, toy, or other household object) may be located. Provide a clear reasoning for your choice, considering visual cues such as doors, furniture, and activity indicators.\\
     
    Input: [RGB Image with Waypoint Marking]\\

    Constraints:
    \begin{itemize}
        \item Evaluate each waypoint marked in the image.
        \item Explicitly explain why the selected waypoint has the highest probability.
        \item Keep reasoning concise (2-3 sentences max).
        \item Do not suggest waypoints not marked in the image.
    \end{itemize}
    
    Model Output: JSON List \\
{[} \\
``Selected waypoint'': ``P2'', \\
``Reasoning'': ``P2 has a door leading to a room where children's activity is visible, making it the most likely location for the target object. P1 leads to a bathroom with minimal activity, and P3 is the kitchen area where children are unlikely to play.'' \\
{]}

\end{tcolorbox}

\subsubsection{Recovery strategy Prompts}

\begin{tcolorbox}[
    coltitle=white,         
    title=\large{\scriptsize Recovery strategy Prompts}, 
    arc=0mm,                
    boxrule=0mm,            
    toptitle=1mm,           
    bottomtitle=1mm,        
    left=4mm,               
    right=4mm,              
    top=2mm,                
    bottom=2mm,             
    breakable,
    fontupper=\itshape\scriptsize     
]
Task: You are the helper for fine-tuning the robot's movements. The robot has already reached the selected KeyPoint. Generate a sequence of fine-grained navigation actions to approach the target area (e.g., stairs or object) from this KeyPoint. Provide reasoning for each action based on visual cues and local map layout, considering obstacles, doorways, and alignment with the target.

Inputs: \\
1. Visual Input: [Single RGB Image at current KeyPoint] \\
2. Local Map: [Top-down view showing robot position, obstacles, and target area]\\

Constraints:
\begin{itemize}
    \item Select from the actions list \{\text{Move Forward}, \text{Turn Left}, \text{Turn Right}, \text{Look Up}, \text{Look Down}, \text{Stop}\}.
    \item Provide reasoning for every action in 1-2 sentences.
    \item Combine RGB observation and map information for decision-making.
    \item Ensure the robot avoids obstacles and aligns efficiently toward the stairs or target area.
\end{itemize}

Model Output: JSON List \\
{[} \\
``Actions'': ``$[$ Go Forward, Go Forward, Turn Right, Turn Right $]$'', \\
``Reasoning'': ``
    S1: Go Forward – The current view faces the window with stairs to the right, moving forward leaves the window area and approaches the staircase base; \\
    S2: Go Forward – Continue forward to position within the reachable staircase area; \\
    S3: Turn Right – Adjust orientation toward the staircase entrance so steps begin to appear; \\
    S4: Turn Right – Fine-tune the view, fully revealing the staircase and allowing direct approach to the entrance.
'' \\
{]}
\end{tcolorbox}

\subsubsection{Reminiscing strategy Prompts}

\begin{tcolorbox}[
    coltitle=white,
    title=\large{\scriptsize Reminiscing strategy Prompts (Before reaching the selected KeyPoint)},
    arc=0mm,
    boxrule=0mm,
    toptitle=1mm,
    bottomtitle=1mm,
    left=4mm,
    right=4mm,
    top=2mm,
    bottom=2mm,
    breakable,
    fontupper={\itshape\scriptsize}
]

Task: You are a  Reminiscing Assistant. Analyze a set of RGB images and a top-down map with waypoint annotations. Your goal is to identify: 
1. Which image(s) may contain a forgotten or overlooked target object.  
2. Which image(s) may indicate a potential but previously unobserved stair entrance.  

For each identification, provide a concise reasoning based on visual cues (furniture, doors, objects) and map context (edges of rooms, hallway junctions, corners).\\

Inputs: \\
1. RGB Image Set: [Rgb Images 1–n] \\
2. KeyPoint Map: [Top-down view with Keypoints, robot trajectory, and explored areas] \\

Constraints:
\begin{itemize}
    \item Only select from the provided RGB images.
    \item Provide reasoning for each selection in 1-2 sentences.
    \item Use map context to infer unexplored or partially observed areas.
    \item Distinguish between probable target object locations and potential stair entrances.
\end{itemize}

Model Output: JSON List \\
{[} \\
``Potential Target Object'': ``Image 2'', \\
``Target Reasoning'': ``Image 2 shows a doorway to a bedroom with visible toys or activity, making it the most likely location for a forgotten target object.'', \\
``Potential Stair Entrance'': ``Image 1'', \\
``Stair Reasoning'': ``Image 1 is at the edge of the house and at a junction between floors, which is typical for a stair location. No other images indicate vertical access points.''\\
{]}

\end{tcolorbox}

\begin{tcolorbox}[
    coltitle=white,
    title=\large{\scriptsize Reminiscing strategy Prompts (After reaching the selected KeyPoint)}, 
    arc=0mm,
    boxrule=0mm,
    toptitle=1mm,
    bottomtitle=1mm,
    left=4mm,
    right=4mm,
    top=2mm,
    bottom=2mm,
    breakable,
    fontupper={\itshape\scriptsize}
]

Task: You are the helper for fine-tuning the robot's movements. The robot has already reached the selected KeyPoint. Generate a sequence of fine-grained navigation actions to approach the target area (e.g., stairs or object) from this KeyPoint. Provide reasoning for each action based on visual cues and local map layout, considering obstacles, doorways, and alignment with the target.

Inputs: \\
1. Visual Input: [Single RGB Image at current KeyPoint] \\
2. Local Map: [Top-down view showing robot position, obstacles, and target area]\\

Constraints:
\begin{itemize}
    \item Select from the actions list \{\text{Move Forward}, \text{Turn Left}, \text{Turn Right}, \text{Look Up}, \text{Look Down}, \text{Stop}\}.
    \item Provide reasoning for every action in 1-2 sentences.
    \item Combine RGB observation and map information for decision-making.
    \item Ensure the robot avoids obstacles and aligns efficiently toward the stairs or target area.
\end{itemize}

Model Output: JSON List \\
{[} \\
``Actions'': ``$[$ Go Forward, Go Forward, Turn Right, Turn Right $]$'', \\
``Reasoning'': ``
    S1: Go Forward – The current view faces the window with stairs to the right, moving forward leaves the window area and approaches the staircase base; \\
    S2: Go Forward – Continue forward to position within the reachable staircase area; \\
    S3: Turn Right – Adjust orientation toward the staircase entrance so steps begin to appear; \\
    S4: Turn Right – Fine-tune the view, fully revealing the staircase and allowing direct approach to the entrance.
'' \\
{]}
\end{tcolorbox}

\subsection{Experimental Details}
\label{sec:EDE}

\subsubsection{Experimental Parameters.} We will provide some specific parameters for the initialization of the experiment. In adaptive exploration strategy, the fixed weights $\alpha$ and $\beta$ in the value function $V(f)$ are initialized to $0.5$, $d_\text{max}$ in $S_{\text{dist}}(f)$ is initialized to 20, which is set to the maximum distance the robot can travel, $\alpha(k)$ and $\beta(k)$, which are factors in the optimization function $J(f)$, are initialized to $0.5$ and vary with the step index $k$. $\sigma_1$, $\sigma_2$, and $\sigma_3$, which are coefficients in the exploration reward $ER(k)$, are  initialized to 2, 1 and 1. $K_\text{max}$ and $N_\text{total}$ are set to 500 and 50. In strategy transition, we set $N_\text{rec}$, $SN_\text{rec}$ and $D_\text{rec}$ to 10, 5 and 0.5.\\

\subsubsection{Experimental Ablative Study Details}
\label{sec:Experimental Ablative Study Details}

\noindent \textbf{Impact of  Parameters in Value-Function-Guided Exploration Strategy}. As shown in Table~\ref{tab:ablation2}, combining semantic
and uncertainty cues with equal weights (c) outperforms
either component alone (a, b) on both datasets.
Compared with semantic-only exploration, the combined
configuration improves SR and SPL by 1.3 and 4.1
percentage points on HM3D, and by 2.2 and 1.5 points
on MP3D, respectively.
These results suggest that the two cues provide
complementary information for exploration.
Replacing equal static weights with dynamic weighting
(c $\rightarrow$ e) further increases SR/SPL by 0.8/0.4
points on HM3D and 2.6/1.0 points on MP3D, indicating
that adjusting their relative contributions benefits
both navigation success and path efficiency.
Configuration (f) achieves the highest SR and SPL
on both benchmarks.

\begin{table*}[t]
    \centering
    \caption{Ablation of value-function-guided exploration
    on HM3D and MP3D, comparing semantic and uncertainty
    components and static versus dynamic weighting.}
    \label{tab:ablation2}
    \setlength{\tabcolsep}{5pt}
    \begin{tabular}{l cc cc}
        \toprule
        \multirow{2}{*}{Exploration Strategy}
        & \multicolumn{2}{c}{HM3D}
        & \multicolumn{2}{c}{MP3D} \\
        \cmidrule(lr){2-3} \cmidrule(lr){4-5}
        & SR$\uparrow$ & SPL$\uparrow$
        & SR$\uparrow$ & SPL$\uparrow$ \\
        \midrule
        (a) Semantic ($\alpha=1, \beta=0$)
        & 67.2 & 29.3 & 41.1 & 14.8 \\
        (b) Uncertainty ($\alpha=0, \beta=1$)
        & 66.8 & 31.1 & 41.8 & 15.5 \\
        (c) Semantic + Uncertainty ($\alpha=0.5, \beta=0.5$)
        & 68.5 & 33.4 & 43.3 & 16.3 \\
        (d) Semantic + Uncertainty ($\alpha=0.5, \beta=0.5$)
        & 69.8 & 34.2 & 45.4 & 17.1 \\
        (e) Semantic + Uncertainty (Dynamic $\alpha, \beta$)
        & 69.3 & 33.8 & 45.9 & 17.3 \\
        \textbf{(f) Semantic + Uncertainty}
        (Dynamic $\alpha, \beta$)
        & \textbf{72.3} & \textbf{35.6}
        & \textbf{47.9} & \textbf{18.7} \\
        \bottomrule
    \end{tabular}
\end{table*}

\noindent \textbf{Effectiveness of VLM and LLM for Navigation Strategy}. As shown in Tab. \ref{tab:ablation3}, using Gemini results in a baseline SR of 70.8\% and  SPL of 33.8\%. The Qwen2.5-VL model performs slightly better, achieving an SR of 71.4\% and an SPL of 34.5\%. However, GPT-4o demonstrates the highest efficacy. This indicates that GPT-4o provides superior spatial reasoning and zero-shot generalization capabilities. As shown in Tab. \ref{tab:ablation4},  CLIP\cite{Clip} and BLIP-2\cite{Blip-2} achieve comparable performance—yielding SRs of 72.1\% and 72.3\%, and SPLs of 35.5\% and 35.6\%, respectively—both of which outperform the BLIP baseline. Notably, BLIP2 maintains a slight edge over CLIP in zero-shot text-to-image retrieval accuracy.\\

\begin{table}[htbp]
    \centering
    
    \begin{minipage}{0.48\textwidth}
        \centering
        \caption{Effect of the LLM for Navigation Strategy on HM3D.}
        \label{tab:ablation3}
        \setlength{\tabcolsep}{5pt}
        \begin{tabular}{ccc}
        \toprule
        Model & SR$\uparrow$ & SPL$\uparrow$ \\
        \midrule
        Gemini       & 70.8 & 33.8 \\
        Qwen2.5-VL & 71.4 & 34.5 \\
        \textbf{GPT-4o}  & \textbf{72.3} & \textbf{35.6} \\
        \bottomrule
        \end{tabular}
    \end{minipage}\hfill 
    \begin{minipage}{0.48\textwidth}
        \centering
        \caption{Effect of the VLM for Navigation Strategy on HM3D.}
        \label{tab:ablation4}
        \setlength{\tabcolsep}{5pt}
        \begin{tabular}{ccc}
        \toprule
        Model & SR$\uparrow$ & SPL$\uparrow$ \\
        \midrule
        Blip\cite{blip}       & 70.9 & 35.1 \\
        Clip\cite{Clip} & 72.1 & 35.5 \\
        \textbf{Blip-2}\cite{Blip-2}  & \textbf{72.3} & \textbf{35.6} \\
        \bottomrule
        \end{tabular}
    \end{minipage}
    
\end{table}

\subsubsection{Experimental Demo Details}
\label{sec:Experimental Demo Details}

Please refer to our supplementary video material for more details. The video is divided into third sections: the first section provides a full walkthrough of two complete navigation episodes using our LERR-Nav; the second section presents additional comparison experiments with ASCENT\cite{Stairway}, highlighting the advantages of our approach in terms of exploration speed and robustness.
The third stage demonstrates our method’s ability to generalize to real-world environments.

\begin{figure}[t]
    \centering
    \includegraphics[width=\textwidth]{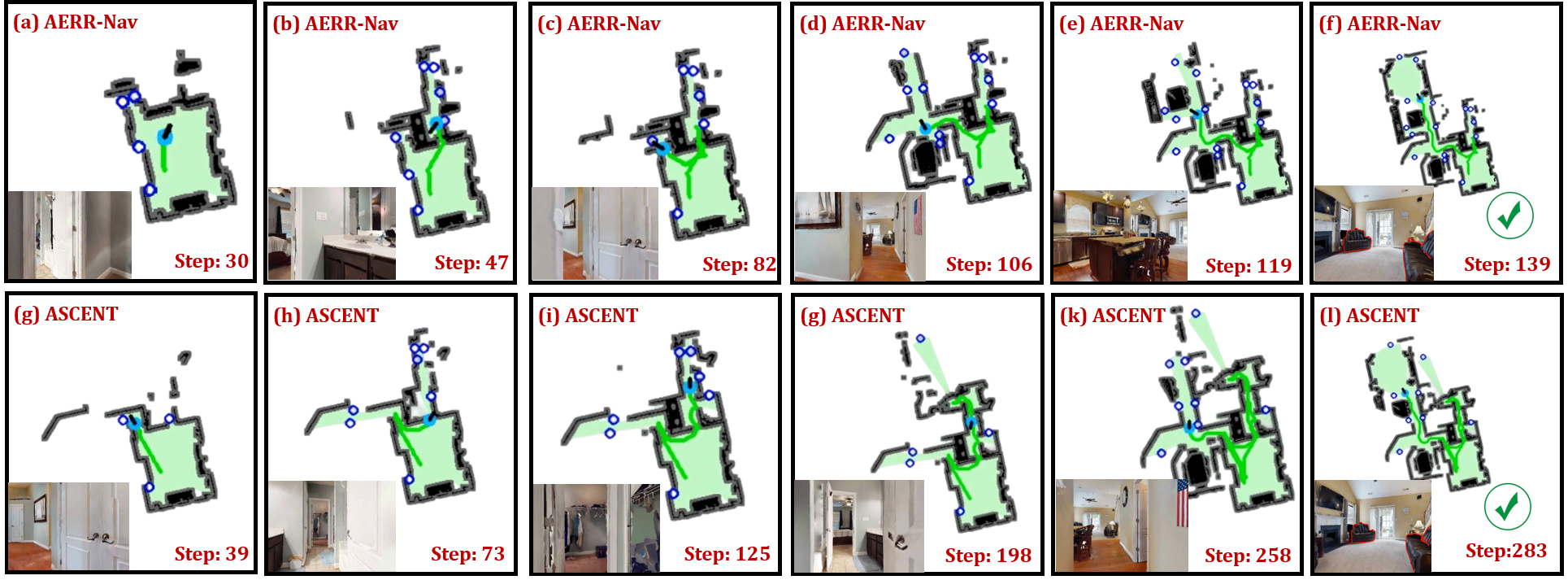}
    \caption{Case Study1: (a)–(f)/(g)–(l) illustrate the complete process of  LERR-Nav/ASCENT\cite{Stairway} searching for the sofa.}
    \label{fig:5}
\end{figure}

As shown in Fig. \ref{fig:5}, our LERR-Nav completes the task with fewer time steps, highlighting the efficiency of our exploration strategy. The underlying reason is as follows. Comparing Fig. \ref{fig:5}(c) and Fig. \ref{fig:5}(i), the left half of the environment is dominated by a corridor, whereas the right half mainly contains a bathroom and a dressing room. Our LERR-Nav chooses to explore the hall, while ASCENT explores the bathroom and dressing room, which results in a substantial waste of steps for ASCENT. ASCENT made this choice because its coarse-grained exploration selects the nearest frontier point, in other words, LLM reasoning is only triggered when no frontier points exist near the robot. In contrast, our LERR-Nav triggers LLM reasoning upon detecting a door. LLM infers that sofa is unlikely in dressing room or bathroom, thus guiding the robot to explore the hallway as shown in Fig. \ref{fig:5}(d). 

\begin{figure}[t]
    \centering
    \includegraphics[width=\textwidth]{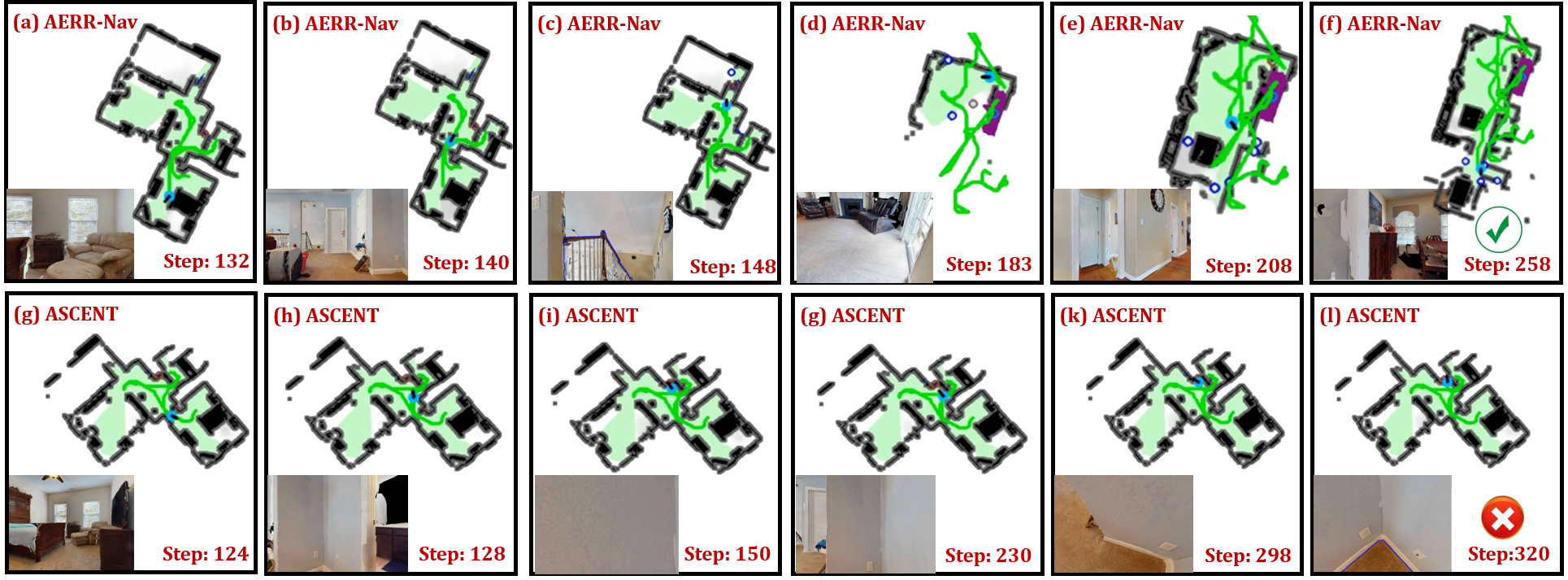}
    \caption{Case Study2: The visualization starts after each method has finished exploring the single-floor. (a)–(f)/(g)–(l) show the process of our LERR-Nav/ASCENT\cite{Stairway} searching for the potted plant.}
    \label{fig:6}
\end{figure}

As shown in Fig. \ref{fig:6}, our LERR-Nav enables the robot to recall and localize the staircase entrance, allowing it to continue and complete the cross-floor task, which demonstrates the importance of the reminiscing strategy.  As illustrated in Fig. \ref{fig:6}(a), when no other frontier points are available, LERR-Nav enters a recall mode. As shown in Fig. \ref{fig:6}(b) and \ref{fig:6}(c), the LLM then guides the agent to hypothesize a likely location of the staircase entrance and successfully discovers the staircase. This ultimately enables the agent to accomplish the cross-floor task. In contrast,  ASCENT fails to identify the staircase entrance and keeps wandering repeatedly, resulting in task failure.

\subsubsection{Supplementary Figures}
\label{sec:Supplementary Figures}
\begin{figure}[t]
    \centering
    \includegraphics[width=\textwidth]{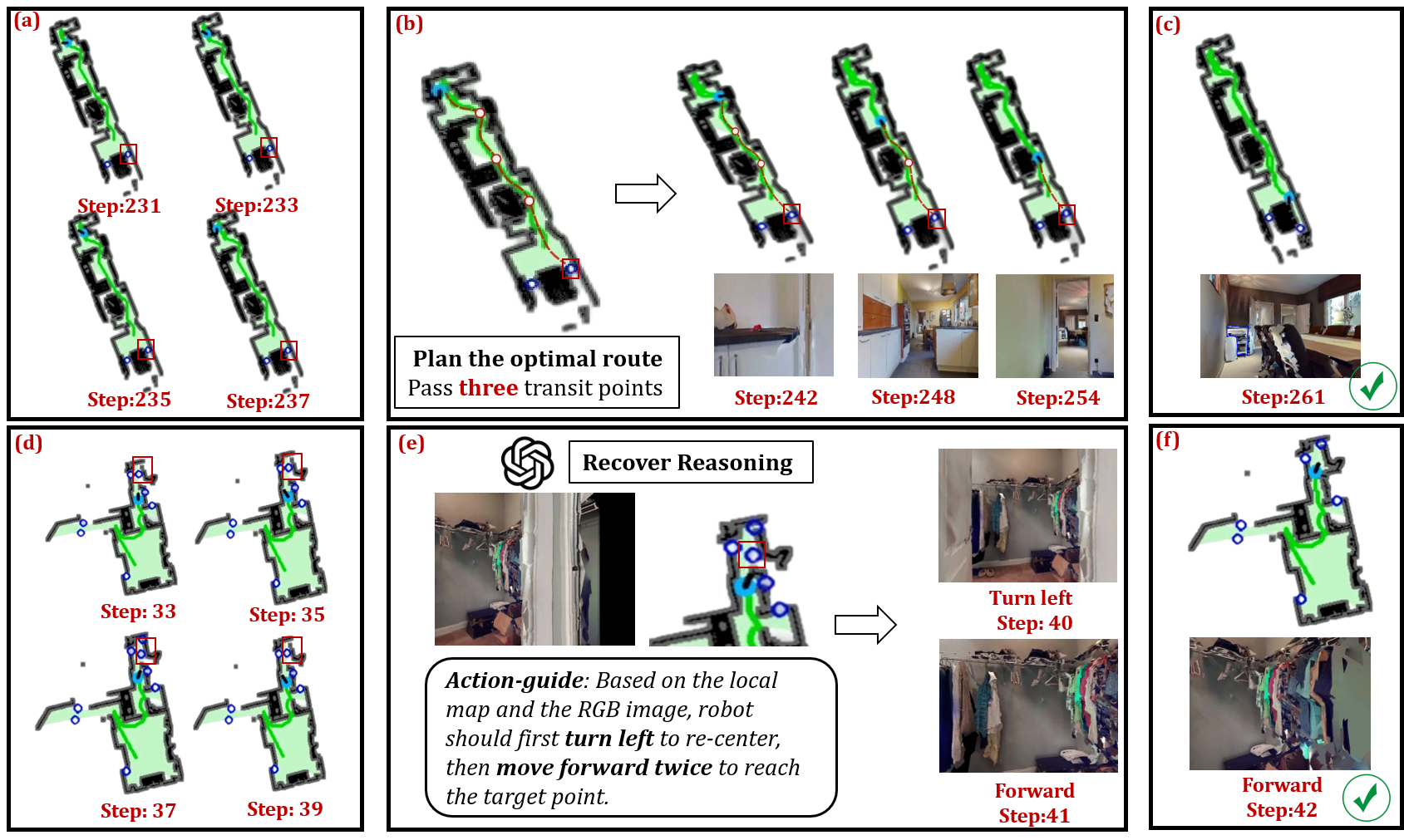}
    \caption{Explanation of the recovery strategy workflow: (a) and (d) illustrate abnormal behaviors caused by frontier points distance. (b)$\rightarrow$(c) demonstrate how to handle overly distant frontier points using a segmented method. (e)$\rightarrow$(f) show how LLM progressively approaches the target frontier point via fine-grained action adjustments.}
    \label{fig:4}
\end{figure}

\begin{figure}[t]
    \centering
    \includegraphics[width=\textwidth]{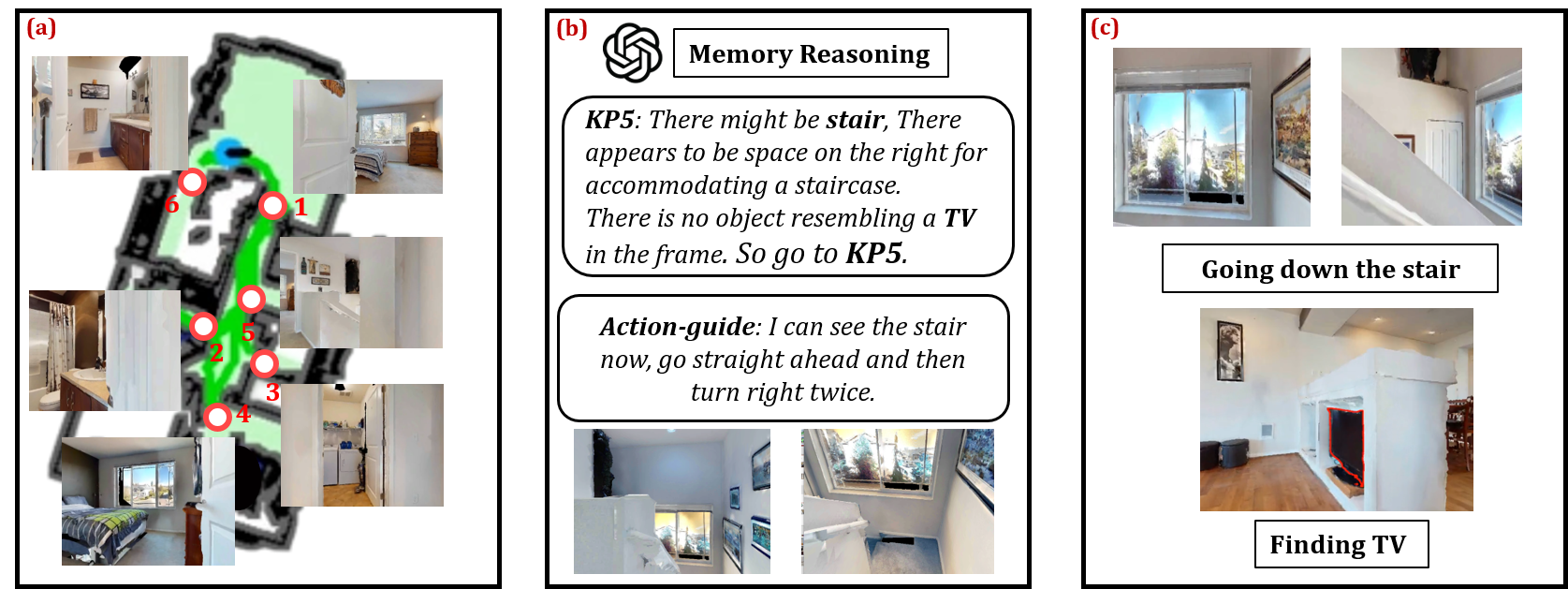}
    \caption{Explanation of how the reminiscing strategy facilitates progress in task: (a) visualizes the  representation of the KeyPoint Map. (b) illustrates how the LLM  chose to keypoints and locate the staircase entrance. (c) shows the robot descending the stairs after locating the stairway entrance and then continuing to find the target.}
    \label{fig:3}
\end{figure}

\end{document}